\documentclass{article}

 \usepackage[preprint]{neurips_2026}

\usepackage[utf8]{inputenc} 
\usepackage[T1]{fontenc}    
\usepackage{hyperref}       
\usepackage{url}            
\usepackage{booktabs}       
\usepackage{amsfonts}       
\usepackage{nicefrac}       
\usepackage{microtype}      
\usepackage{xcolor}  
\usepackage{multirow}
\usepackage{enumitem}
\usepackage{amsmath,amssymb,amsthm}
\usepackage{graphicx}
\usepackage{wrapfig}
\usepackage{subcaption}

\theoremstyle{plain}
\newtheorem{theorem}{Theorem}[section]

\newtheorem{lemma}[theorem]{Lemma}
\newtheorem{corollary}[theorem]{Corollary}
\theoremstyle{definition}
\newtheorem{definition}[theorem]{Definition}

\theoremstyle{remark}

\theoremstyle{problem}
\newtheorem{problem}[theorem]{Problem}

\title{SOMA: Efficient Multi-turn LLM Serving via Small Language Model}

\author{%
\begin{tabular}{c}
\small
\textbf{Xueqi Cheng}$^{1}$ \quad
\textbf{Qiong Wu}$^{2}$ \quad
\textbf{Zhengyi Zhou}$^{2}$ \quad
\textbf{Xugui Zhou}$^{3}$ \quad
\textbf{Tyler Derr}$^{4}$ \quad
\textbf{Yushun Dong}$^{1}$ \\
{\normalfont\footnotesize
$^{1}$Florida State University \quad
$^{2}$AT\&T Chief Data Office \quad
$^{3}$Louisiana State University \quad
$^{4}$Vanderbilt University} \\
{\normalfont\ttfamily\scriptsize
\{xc25,yushun.dong\}@fsu.edu;\ \{qw6547,zz547k\}@att.com} \
{\normalfont\ttfamily\scriptsize
xuguizhou@lsu.edu;\ tyler.derr@vanderbilt.edu}
\end{tabular}%
}

\begin{document}
\maketitle

\begin{abstract}

Large Language Models (LLMs) are increasingly deployed in multi-turn dialogue settings where preserving conversational context across turns is essential. A standard serving practice concatenates the full dialogue history at every turn, which reliably maintains coherence but incurs substantial cost in latency, memory, and API expenditure, especially when queries are routed to large proprietary models. Existing approaches often struggle to balance the trade-off between response quality and efficiency. We propose a framework that exploits the early turns of a session to estimate a local response manifold and then adapt a smaller surrogate model to this local region for the remainder of the conversation. Concretely, we learn soft prompts that maximize semantic divergence between the large and surrogate small language models' responses to surface least-aligned local directions, stabilize training with anti-degeneration control, and distill the mined cases into localized LoRA fine-tuning so the surrogate runs without prompts at inference. A simple gate enables a one-time switch with rollback on drift. We further provide a theoretical analysis for key components in SOMA. Extensive experiments show the effectiveness of SOMA. The source code is provided at: 
\url{https://github.com/LabRAI/SOMA}.
\end{abstract}

\section{Introduction}\label{sec:intro}


Large language models (LLMs) such as the GPT series~\citep{radford2019language, brown2020language, achiam2023gpt}, LLaMA~\citep{touvron2023llama}, Claude~\citep{anthropic_introducing_claude_2023}, and DeepSeek~\citep{guo2025deepseek} have demonstrated strong performance in real-world machine-learning-as-a-service (MLaaS) applications, ranging from chat assistants to code generation~\citep{park2023thinking, dong2023towards, liu2024exploring, li2026llmclinicalgraphstructure, yu2026health}. As LLMs are increasingly deployed in interactive settings, \textit{multi-turn LLM serving}, involving extended interactions between humans and LLMs or among multiple LLM agents, has emerged as a key research focus, as it better reflects real-world usage scenarios~\citep{yi2024survey, li2025beyond, zhao2025amplifying}.  Existing research reveals that multi-turn interactions are widespread, underscoring the need for serving systems capable of handling extended conversations in a context-aware manner~\citep{chen2024sharegpt4v, gao2024cost}.
However, supporting efficient context-dependent 
multi-turn interaction remains a key challenge, as most LLM serving systems are stateless and require resending the entire conversation history, including all prior queries and responses, with each new query to generate a new response~\citep{ananda2025beyond, moon2025accelerating}. This leads to redundant computation, high latency, and rising serving costs as conversations lengthen.

Previous work has explored efficient multi-turn LLM serving through two main approaches. One line of work focuses on \textit{single-model methods} that compress dialogue history~\citep{wang2025recursively, chen2024compress, xiao2024efficient}, retrieve memory from external modules~\citep{melz2023enhancing, gutierrez2024hipporag}, or reuse attention computations~\citep{gao2024cost, jeong2025accelerating, anthropic_prompt_caching_2024}. However, these still rely heavily on large LLMs for every turn, leading to high monetary cost, latency, and GPU usage. They also often truncate or overlook extended context, limiting reasoning over long dialogues. Another line adopts \textit{multi-model methods}, routing simple queries to smaller models while escalating harder ones to larger LLMs~\citep{behera2025towards, schick2023toolformer, ding2024hybrid, shnitzer2023large}. Yet, small models struggle to generalize across dialogue complexity, and model switching introduces additional overhead. Moreover, LLMs are known to rely on early turns~\citep{xiao2024efficient, laban2025llms}, compounding the difficulty of maintaining coherence in multi-turn settings. 

\textit{\textbf{Is it possible to achieve an efficient, context-aware multi-turn LLM serving framework that avoids recomputing the full history at every turn while preserving response quality?}}

To achieve this goal, we perform in-depth explorations of real-world multi-turn dialogues across different domains and reveal an interesting long-tail distribution in token counts across turns: early turns are substantially longer, while later turns gradually decrease in length. This phenomenon aligns with the intuition of prior work that early turns typically carry the substantive openings set issues and anchors, including questions, requests, and proposals, while later turns are typically minimal acknowledgments~\citep{he2018decoupling, stolcke2000dialogue}.
This long-tail trend gives rise to an intuitive idea: since later turns are relatively short, a smaller language model might suffice to generate responses more cost-efficiently. 
However, the bottleneck lies in the “big head”: if the small model must still process the early, information-dense context, the response quality will be degraded. Specifically, while a small language model may behave reasonably at the start, its responses drift from the large model as the dialogue progresses because most grounding is established early, and later turns remain highly dependent on that context. Therefore, simply handing later turns to a small model without modeling the accumulated context degrades quality. To address this, the small language model must not only process shorter inputs but also approximate the larger model’s behavior within the local manifold of its output or hidden space to capture the contextual dependencies shaped by prior dialogue. This defines a local manifold approximation problem, where the small language model aims to replicate the target model’s behavior induced by the current conversational context.

Building on these insights, we propose a novel framework for efficient multi-turn LLM serving that enables a small language model to locally approximate the behavior of a large language model within a constrained region of the reasoning manifold. Specifically, we present SOMA (\textit{S}oft-prompts for l\textit{O}cal \textit{M}anifold \textit{A}pproximation) to dynamically adapt the small language model to the local behavior of the larger model conditioned on early turn interactions. This is achieved through a three-stage pipeline: (1) \textit{Soft prompt tuning}, where we efficiently explore the local reasoning manifold induced by the early conversational context to identify directions of maximal behavioral divergence between the small and large language model
; (2) \textit{Localized fine-tuning}, where we efficiently fine-tune the small language model on a small number of input–output pairs to align it with the larger model within this context-specific region of the manifold; and (3) \textit{Efficiency inference}, where we incorporate the extractive summary to minimize computational overhead and the rollback mechanism that monitor potential topic shift to maintain service quality. Together, these components allow the small model to effectively approximate the larger model’s reasoning process within the context of a given session, enabling both cost-effective and context-aware multi-turn serving. Extensive experiments show the effectiveness of our proposed method. Overall, our contributions are: 
\begin{itemize}[left=0pt,itemsep=2pt]
\item \textbf{Long-tail pattern in multi-turn dialogues.} 
We first reveal a previously under-explored long-tail pattern in multi-turn dialogues: the first few turns concentrate heavy context, while later turns are shorter yet more dependent on previous turns. This key empirical characterization suggests that substantial computational and monetary savings can be achieved if a smaller and cheaper language model can replace a large one to process the later turns when given the accumulated context.



\item \textbf{SOMA: Efficient multi-turn serving.} It first learns soft prompts that expose the largest surrogate–original response dissimilarity, then adapts the surrogate localized LoRA accordingly, enabling prompt-free inference with a simple cosine gate for switching and rollback.

\item \textbf{Theory analysis and empirical evaluation.} 
We provide concentration-based bounds for switching, coverage guarantees for prompt-direction search, and suboptimality limits for selected directions. Guided by these results, empirical studies show the effectiveness of SOMA in real world.
\end{itemize}


\section{Preliminaries}\label{sec:prelim}


\subsection{Notations}

In this paper, a multi-turn dialogue prefix of length $k$ is $\mathcal{D}_k=\{(q_1,a_1),\ldots,(q_k,a_k)\}$ where \( q_t \) is the user query at turn \( t \), and \( a_t \) is the corresponding model response. $F$ represents the original proprietary black-box LLM, which is a common setting in machine-learning-as-a-service (MLaaS), and $G$ is the surrogate small language model. The textual response at turn $t$ is $a_t^M$ for $M\in\{F,G\}$. Let $f_M(\cdot)$ be a feature map to the hidden space, and $\mathbf{h}_t=f_M(q_{\le t})\in\mathbb{R}^d$ the hidden state at turn $t$. The first $k$ hidden states form $\mathcal{H}_k=\{\mathbf{h}_1,\ldots,\mathbf{h}_k\}$ and induce a local manifold $\mathcal{M}_k^M\subset\mathbb{R}^d$. 
A length-$L$ soft prompt is $\mathbf{P}\in\mathbb{R}^{L\times d}$. 
More details are given in Appendix~\ref{sec:notation}.

\subsection{Exploring Token--Turn Patterns in Multi-Turn LLM Dialogues}
\label{sec:long_tail}

\begin{wrapfigure}[15]{r}{0.44\textwidth}
  \vspace{-11pt}
  \centering
  \includegraphics[width=0.439\textwidth]{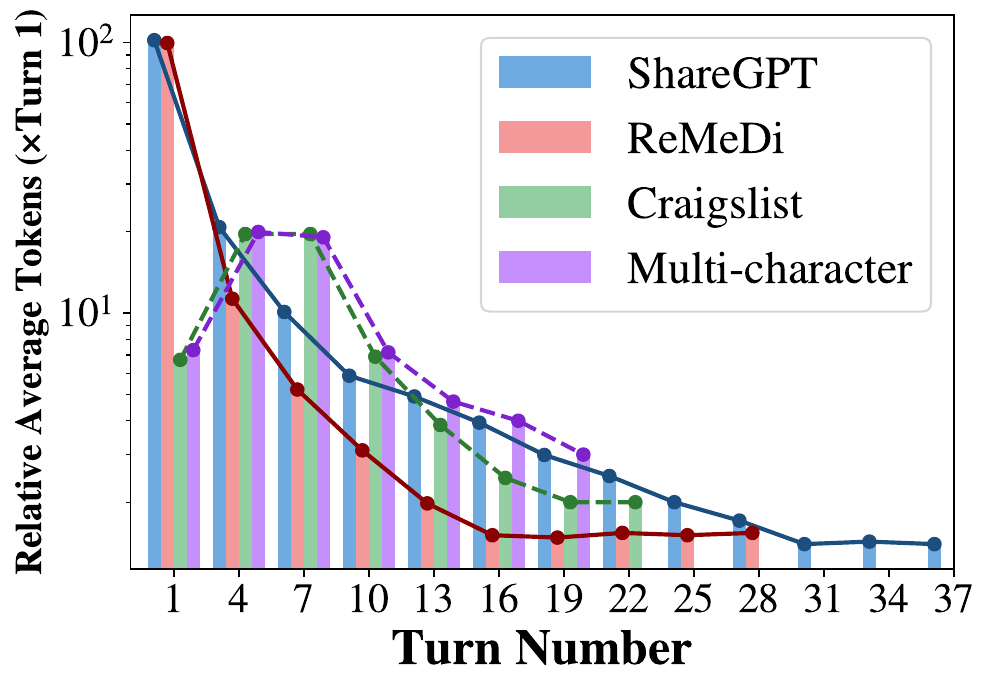}
  \vspace{-18pt}
  \caption{Relative average token count per turn, normalized by Turn 1. Across four dialogue datasets, token usage drops after the early turns and then forms a long tail.}
  \vspace{-11pt}
  \label{fig:longtail-token-decay}
\end{wrapfigure}

Efficient multi-turn serving depends on how the dialogue state evolves over time. In standard LLM serving, each new request is processed together with the full previous history, so the computational cost grows with the length of the conversation. However, the amount of new information introduced at each turn may follow a different pattern. We therefore first study how token usage changes across dialogue turns.

We consider four representative multi-turn settings: \textit{ShareGPT}~\citep{chen2024sharegpt4v} for open-ended human--LLM chat, \textit{ReMeDi}~\citep{yan2022remedi} for doctor--patient consultation, \textit{Craigslist Bargain}~\citep{he2018decoupling} for negotiation, and \textit{Multi-Character}~\citep{agentlans_multi-character_dialogue_2024} for multi-party role-play. These datasets cover different speaker roles, task goals, and interaction structures. For each dataset, we compute the average token count at turn \(t\) and normalize it by the average token count of Turn 1.

Figure~\ref{fig:longtail-token-decay} shows a consistent long-tail pattern. Early turns are token-heavy because they usually introduce the task, topic, constraints, and speaker intent. Later turns are much shorter and stabilize at a lower level. This does not make later turns independent; rather, it suggests that many later turns are interpreted as local updates under an already established dialogue state. As a result, the marginal new input becomes smaller while standard serving still repeatedly processes the full prefix. This mismatch motivates an adaptive serving strategy: use the original large model to establish the early dialogue state, and then use a cheaper surrogate when subsequent turns remain within that local state.

\subsection{Local Manifold Approximation for Multi-Turn Dialogues}

The long-tail pattern suggests an opportunity for surrogate serving, but directly replacing the original model with a smaller model is unreliable. A later turn is often short because much of its meaning is supplied by the previous dialogue. If the surrogate is not aligned with the original model under this prefix, it may produce fluent responses that still drift from the intended context.

We describe this challenge through a local manifold view. Let \(F\) denote the original large model and \(G\) denote the small surrogate. Given a dialogue prefix
\[
\mathcal{D}_{k}=\{(q_1,a_1),\ldots,(q_k,a_k)\},
\]
the prefix induces a local region in the model's contextual representation space. This region captures the current topic, constraints, style, and task state of the conversation. It adapts \(G\) to match the local behavior of \(F\) around the current dialogue prefix.

\begin{problem}[Local Manifold Approximation for Multi-Turn Interactions]
Given a dialogue prefix \(\mathcal{D}_{k}\) and an original model \(F\), learn a surrogate model \(G\in\mathcal{G}\) whose local representation region matches that of \(F\) under the same prefix:
\[
\min_{G\in\mathcal{G}}
\mathrm{dist}\!\left(
\mathcal{M}^{G}_{k}(\mathcal{D}_{k}),
\mathcal{M}^{F}_{k}(\mathcal{D}_{k})
\right),
\]
where \(\mathcal{M}^{F}_{k}(\mathcal{D}_{k})\) and \(\mathcal{M}^{G}_{k}(\mathcal{D}_{k})\) denote the local representation regions induced by the prefix, and \(\mathrm{dist}(\cdot,\cdot)\) measures their discrepancy.
\end{problem}

This formulation leads to the design of SOMA. The early turns are used to expose where the surrogate disagrees with the original model within the current dialogue state. These locally informative examples are then used to adapt the surrogate. After adaptation, the surrogate serves later turns only while the dialogue remains close to the learned local region; when the conversation drifts, the system returns to the original model and refreshes the local state.

\section{SOMA: Local Manifold Approximation via Soft Prompts}\label{sec:method}

SOMA combines three components to enable efficient local serving. First, the original model \(F\) handles early turns to establish the dialogue state. Second, soft-prompt mining identifies where the surrogate \(G\) is least aligned with \(F\), and these hard local cases are distilled into a lightweight LoRA adapter. Third, a semantic gate switches to \(G\) only when local alignment is sufficient, while a drift detector rolls back to \(F\) when the dialogue moves away from the learned local state.

\subsection{Initialization}

We initialize a soft prompt matrix $\mathbf{P}\in\mathbb{R}^{L_p\times d}$ on the surrogate $G$. Each row is sampled i.i.d. from $\boldsymbol{p}_{\ell}\sim\mathcal{N}(\boldsymbol{0},\sigma^2\mathbf{I}_d)$, where $\sigma>0$ is the initialization scale. Let $\mathbf{E}=[\boldsymbol e_v]_{v\in\mathcal V}\in\mathbb{R}^{d\times |\mathcal V|}$ be the surrogate embedding matrix and let $\mathrm{tok}(v)$ denote the token string associated with index $v$.

Because the original model $F$ cannot accept continuous embedding vectors, the soft prompt is used in two aligned forms. The surrogate receives $\mathbf{P}$ directly at the embedding layer, while the original model receives a nearest-neighbor verbalization of the same prompt:
\[
v_{\ell}=\arg\max_{v\in\mathcal V}
\frac{\langle \boldsymbol p_{\ell},\boldsymbol e_v\rangle}{\|\boldsymbol p_{\ell}\|_2\|\boldsymbol e_v\|_2},
\qquad
V(\mathbf P)=\big(\mathrm{tok}(v_1),\ldots,\mathrm{tok}(v_{L_p})\big).
\]
Let $\mathcal D_{t-1}$ denote the text history before turn $t$ and let $q_t$ be the user query. During prompt mining, the two models are probed under comparable text conditioning:
\[
a_t^{F}=F\!\left(V(\mathbf P)\oplus\mathcal D_{t-1}\oplus q_t\right),
\qquad
 a_t^{G}=G\!\left(\mathbf P\oplus_{\mathrm{emb}}
\mathbf E(\mathrm{tok}_G(\mathcal D_{t-1}\oplus q_t))\right).
\]
Here $\mathrm{tok}_G(\cdot)$ is the tokenizer of $G$, $\mathbf E(\cdot)$ maps text tokens to embeddings, and $\oplus_{\mathrm{emb}}$ concatenates along the sequence dimension. This construction lets $F$ and $G$ be compared under the same dialogue state while allowing gradients to flow only through $\mathbf P$ on $G$.

\subsection{Soft Prompt Tuning for Mining Weak Alignment Directions}

We learn $\mathbf P$ by minimizing a differentiable objective that makes $G$ move away from $F$ on the local dialogue context. The objective has three components: a token-level semantic divergence loss, an expectation-weighted distribution-level divergence term, and an anti-degeneration regularizer.

\textbf{Semantic divergence loss.}
We use an unlikelihood-style loss~\citep{welleck2019neural} to discourage $G$ from assigning high probability to the tokens produced by $F$. A purely exact-token penalty is too narrow, since the surrogate could avoid the teacher token while still assigning high probability to close paraphrases. We therefore define a semantic neighborhood in the surrogate embedding space.

\begin{definition}[Semantic Neighborhood]
For token $u\in\mathcal V$, let $\mathcal N_k(u)$ be the set of the $k$ tokens in $\mathcal V\setminus\{u\}$ whose embeddings have the largest cosine similarity to $\boldsymbol e_u$.
\end{definition}

At turn $t$, the original model produces text $a_t^F$. We tokenize this text using the surrogate tokenizer,
$\boldsymbol y_t^F=\mathrm{tok}_G(a_t^F)=(y_{t,1}^F,\ldots,y_{t,T_F(t)}^F)$, so that the teacher output and surrogate distribution are expressed in the same vocabulary. For each teacher token, we assign temperature-weighted mass to the token and its embedding neighbors:
\[
s_{\tau}(v\mid y_{t,i}^F)=
\frac{\exp(\cos(\boldsymbol e_v,\boldsymbol e_{y_{t,i}^F})/\tau)}
{\sum_{u\in\{y_{t,i}^F\}\cup\mathcal N_k(y_{t,i}^F)}
\exp(\cos(\boldsymbol e_u,\boldsymbol e_{y_{t,i}^F})/\tau)},
\qquad \tau>0 .
\]
Let $S_t=\mathcal D_{t-1}\oplus q_t$. With causal masking, a single forward pass of $G$ on the prefix $S_t$ and the teacher answer prefix gives the next-token distributions
$\{\Pi_{t,i}(\mathbf P)\}_{i=1}^{T_F(t)}$, where
$\Pi_{t,i}(\mathbf P)=\pi_G(\cdot\mid S_t;\mathbf P, a_{t,<i}^F)$.
The semantic divergence loss is
\begin{equation}
\label{eq:sem-with-aF}
\mathcal L_{\mathrm{sem}}(\mathbf P;\mathcal D_{t-1},q_t)=
\frac{1}{T_F(t)}\sum_{i=1}^{T_F(t)}
\sum_{v\in\{y_{t,i}^F\}\cup\mathcal N_k(y_{t,i}^F)}
s_{\tau}(v\mid y_{t,i}^F)
\Big[-\log\big(1-\Pi_{t,i}(\mathbf P)[v]\big)\Big].
\end{equation}
Minimizing Eq.~\ref{eq:sem-with-aF} searches for prompts that expose local disagreement between $F$ and $G$ not only on exact tokens but also on nearby semantic alternatives.

\textbf{Expectation-weighted semantic divergence.}
The token-neighborhood loss may still miss distribution-level alignment: $G$ can reduce probability on a teacher token and its nearest neighbors while spreading mass over many other tokens with similar meaning. To detect this case, we compare the expected embedding of the surrogate's next-token distribution with the embedding of the teacher token. Since $\mathbf E\in\mathbb R^{d\times |\mathcal V|}$ stores token embeddings as columns, the expected embedding at position $i$ is $\bar{\boldsymbol e}_{t,i}=\mathbf E\Pi_{t,i}(\mathbf P)
=\sum_{v\in\mathcal V}\Pi_{t,i}(\mathbf P)[v] \boldsymbol e_v $. We upweight the neighborhood penalty when this expected embedding still points toward the teacher token $w_{t,i}(\mathbf P)=1+\lambda_{\mathrm{exp}}\,
\mathrm{clip}\!\left(
\cos(\bar{\boldsymbol e}_{t,i},\boldsymbol e_{y_{t,i}^{F}}),0,1
\right)$, where $\lambda_{\mathrm{exp}}\ge 0 $. The final semantic mining loss is then:
\begin{equation}
\label{eq:semexp-with-aF}
\mathcal L_{\mathrm{sem\_exp}}(\mathbf P;\mathcal D_{t-1},q_t)=
\frac{1}{T_F(t)}\sum_{i=1}^{T_F(t)}
w_{t,i}(\mathbf P)
\sum_{v\in\{y_{t,i}^F\}\cup\mathcal N_k(y_{t,i}^F)}
s_{\tau}(v\mid y_{t,i}^{F})
\Big[-\log\big(1-\Pi_{t,i}(\mathbf P)[v]\big)\Big].
\end{equation}
Thus the loss blocks local copying through the neighborhood term and detects broader semantic shadowing through the expectation weight.

\begin{theorem}[Directional Recovery]
\label{thm:dir_recovery}
Let \(\mathbf C=\mathbb E[\mathbf J^\top \mathbf J]\) be the local discrepancy Fisher matrix of \(G\) over the warm-start window, where \(\mathbf J\) is the Jacobian of \(G\)'s log next-token probabilities with respect to the soft prompt \(\mathbf P\). If the local discrepancy is smooth and approximately rank \(r\), then optimizing Eq.~\ref{eq:semexp-with-aF} recovers prompt directions whose span covers at least a \((1-\varepsilon)\) fraction of the top-\(r\) eigenmass of \(\mathbf C\), with smaller \(\varepsilon\) under better warm-start and neighborhood coverage.
\end{theorem}

\textbf{Anti-degeneration regularizer.}
Soft-prompt mining can collapse if $G$ assigns excessive probability to a few high-frequency tokens, producing repetitive or bland continuations that are uninformative about where $F$ and $G$ differ~\citep{li2023repetition,holtzman2019curious,meister2023locally}. We therefore preserve entropy near the prompt-context boundary. Reusing the logits from the same forward pass, define
\[
H_{\mathrm{tail}}(\mathbf P;t)=
\frac{1}{K}\sum_{i\in\mathrm{tail}(t,K)}
\left[-\sum_{v\in\mathcal V}\Pi_{t,i}(\mathbf P)[v]
\log \Pi_{t,i}(\mathbf P)[v]\right].
\]
The complete prompt-mining objective for a minibatch of warm-start turns $\mathcal B$ is
\begin{equation}
\label{eq:prompt-final}
\mathcal J(\mathbf P)=
\frac{1}{|\mathcal B|}\sum_{t\in\mathcal B}
\left[
\mathcal L_{\mathrm{sem\_exp}}(\mathbf P;\mathcal D_{t-1},q_t)
-\beta H_{\mathrm{tail}}(\mathbf P;t)
\right]
+\lambda_P\|\mathbf P\|_F^2 .
\end{equation}
We minimize Eq.~\ref{eq:prompt-final} with AdamW while freezing all weights of $G$. Gradient clipping stabilizes the prompt scale. For efficiency, semantic neighbors are retrieved using an ANN index over normalized token embeddings, and $\bar{\boldsymbol e}_{t,i}$ is computed with a top-$m$ truncation of $\Pi_{t,i}(\mathbf P)$ rather than a dense sum over the full vocabulary. The per-step overhead is therefore linear in the number of teacher-prefix positions and scales with $k+m$ rather than $|\mathcal V|$.

\subsection{Localized Fine-Tuning, Switching, and Rollback}

The learned soft prompts are session-local mining tools. After mining, SOMA converts them into ordinary local supervision for LoRA adaptation. For each warm-start turn, we score its hardness by the largest mined divergence $r_t=\max_{\mathbf P\in\mathcal P_{\mathrm{cand}}}
\mathcal L_{\mathrm{sem\_exp}}(\mathbf P;\mathcal D_{t-1},q_t)$ and form a local training set $\mathcal H$ from the highest-scoring turns and their original-model responses. This makes the data-mining role of soft prompts explicit: Random-FT trains on randomly selected local turns, whereas SOMA trains on turns selected because prompt mining exposes weak alignment directions.

We freeze the base weights of $G$, attach LoRA~\citep{hu2022lora} to attention and MLP projections, and optimize
\begin{equation}
\label{eq:lora-ft}
\mathcal L_{\mathrm{FT}}(\boldsymbol\Theta_{\mathrm{LoRA}})=
\frac{1}{|\mathcal H|}\sum_{(S_t,a_t^F)\in\mathcal H}
\omega_t\,\mathrm{NLL}\big(a_t^F\mid G(S_t;\boldsymbol\Theta_{\mathrm{LoRA}})\big),
\qquad
\omega_t\propto r_t .
\end{equation}
The continuous soft prompts are discarded after this stage. Thus, once SOMA switches, the online serving path uses only the LoRA-adapted surrogate and a compressed text context.

Before switching, SOMA verifies that the adapted surrogate \(G\) is both output-aligned and context-local: it requires high response similarity to the original model \(F\) on a small acceptance batch and checks that the current query remains close to the warm-start dialogue centroid. Once accepted, SOMA serves later turns using a compressed input \(\widetilde S_t\), which keeps a fixed-budget summary of earlier content together with the most recent \(K\) turns, avoiding repeated full-history processing. During surrogate serving, SOMA continuously monitors semantic drift; if recent queries move away from the learned local state, it rolls back to \(F\), refreshes the local summary and centroid, and switches back to \(G\) only after the acceptance gate passes again.
\section{Theoretical Analysis}\label{sec:theory}

This section analyzes three design questions in SOMA: when to accept the session-local surrogate, when warm-start overhead is amortized, and how many soft-prompt candidates are needed. We use the local manifold view only as a first-order approximation around a fixed dialogue state, where small embedding perturbations smoothly change the surrogate's output behavior. SOMA is effective when later turns remain tied to this warm-start state, not because they are intrinsically easy.

\subsection{Warm-Start Reliability and Switching}\label{sec:warm-start}

At turn $t$, let $S_t=\mathcal D_{t-1}\oplus q_t$ be the shared text context and let $a_t^F,a_t^G$ be the outputs of the original model and the LoRA-adapted surrogate. Here we define a bounded discrepancy score
\[
\mathsf{Gap}_{\Theta}(S_t)=1-\cos\big(\phi(a_t^G),\phi(a_t^F)\big)\in[0,1],
\]
where $\Theta$ denotes the current surrogate adapter. A lower value means that the surrogate better preserves the original model's local behavior.

\begin{lemma}[Warm-Start Generalization]\label{lem:warmstart}
Let $h(S)\in[0,1]$ be any bounded local statistic, such as $\mathsf{Gap}_{\Theta}(S)$ or a query-centroid similarity score. Suppose the warm-start stream has effective sample size $W_{\mathrm{eff}}$, accounting for temporal dependence. Then with probability at least $1-\delta$,
\[
\left|\frac{1}{W}\sum_{t=1}^{W}h(S_t)-\mathbb E_{S\sim\mathcal Q}[h(S)]\right|
\le
\sqrt{\frac{2\log(2/\delta)}{W_{\mathrm{eff}}}} .
\]
\end{lemma}

The effective size $W_{\mathrm{eff}}$ decreases when the warm-start window is noisy, heterogeneous, or topic-mixed. Thus the bound predicts the desired behavior: SOMA should require a longer warm start or a more conservative gate when early turns do not define a coherent local state.

Let $\Theta_{\mathrm{old}}$ and $\Theta_{\mathrm{new}}$ denote the surrogate before and after localized adaptation. On an acceptance batch $B$, we define $X_j=\mathsf{Gap}_{\Theta_{\mathrm{old}}}(S_j)-\mathsf{Gap}_{\Theta_{\mathrm{new}}}(S_j)$, then $
\widehat\Delta_B=\frac{1}{|B|}\sum_{S_j\in B}X_j$, where positive $\widehat\Delta_B$ means that adaptation reduces the teacher-surrogate gap.

\begin{theorem}[Acceptance Bound for Switching]\label{thm:detection}
Assume $X_j\in[-1,1]$ and the acceptance batch has effective size $|B|_{\mathrm{eff}}$. Let $\gamma_B(\delta)=\sqrt{\frac{2\log(1/\delta)}{|B|_{\mathrm{eff}}}}$. Then, with probability at least $1-\delta$, $\mathbb E[X]\ge \widehat\Delta_B-\gamma_B(\delta)$. Consequently, if $\widehat\Delta_B\ge \varepsilon+\gamma_B(\delta)$, then the true expected improvement is at least $\varepsilon$ with confidence $1-\delta$.
\end{theorem}

\begin{corollary}[Switching Rule]\label{cor:switch}
Choose $W$ so that the warm-start uncertainty in Lemma~\ref{lem:warmstart} is at most $\eta$, and choose $|B|_{\mathrm{eff}}$ so that $\gamma_B(\delta)\le \gamma$. Switching is justified when the adapted surrogate passes both the output-fidelity gate and the locality gate with margins larger than $\eta+\gamma$. The combined decision error is controlled at confidence at least $1-2\delta$ by a union bound.
\end{corollary}

This result formalizes the intended operating regime. If later turns are short but highly context-dependent, the locality gate can still pass because the turns remain close to the established dialogue state. If the user abruptly changes topic, the locality statistic shifts, the gate fails, and SOMA rolls back to the original model.

\subsection{When Does SOMA Give Net Efficiency Gains?}\label{sec:theory-cost}

SOMA introduces one-time warm-start overhead, so its efficiency should be evaluated at the session level rather than per turn. Let $T$ be the total number of turns and $W$ the switch point. If all turns are served by the original model, the total cost is $C_F(T)=\sum_{t=1}^{T}c_F(\mathcal D_{t-1}\oplus q_t)$, where $c_F(\cdot)$ can denote latency, tokens, or monetary cost. Under SOMA,
\begin{equation}
\label{eq:cost-soma}
\begin{aligned}
C_{\mathrm{SOMA}}(T;W)=
&\sum_{t=1}^{W}c_F(\mathcal D_{t-1}\oplus q_t)
+C_{\mathrm{probe}}(W)+C_{\mathrm{LoRA}}(W) \\
&+\sum_{t=W+1}^{T}c_G(\widetilde S_t)
+C_{\mathrm{rb}}(T;W),
\end{aligned}
\end{equation}
where $C_{\mathrm{probe}}$ is the cost of soft-prompt mining, $C_{\mathrm{LoRA}}$ is the localized adaptation cost, $\widetilde S_t$ is the compressed post-switch input, and $C_{\mathrm{rb}}$ covers drift checks and occasional rollback.

The net gain is
\begin{equation}
\label{eq:net-gain}
\begin{aligned}
\Delta C(T;W)&=C_F(T)-C_{\mathrm{SOMA}}(T;W)\\
&=\sum_{t=W+1}^{T}\left[c_F(\mathcal D_{t-1}\oplus q_t)-c_G(\widetilde S_t)\right]
-C_{\mathrm{probe}}(W)-C_{\mathrm{LoRA}}(W)-C_{\mathrm{rb}}(T;W).
\end{aligned}
\end{equation}
Therefore SOMA is beneficial only when the accumulated post-switch savings exceed the one-time warm-start and adaptation overhead. This explains why very short sessions may not benefit, while medium-to-long locally coherent sessions can amortize the initialization cost.

\subsection{How Many Soft-Prompt Candidates per Iteration?}

Assume that, near a reference prompt $\mathbf P_0$, the prompt-mining objective admits a local quadratic approximation with positive semidefinite curvature $\mathbf H_T$. Let most of its energy lie in an $r_{\mathrm{act}}$-dimensional active subspace, and let $\mathbf u_1$ be the dominant unit direction in that subspace. SOMA samples $M$ prompt candidates and keeps the one with largest mining loss.

\begin{theorem}[Coverage of the Best Local Direction]\label{thm:coverage}
For $r_{\mathrm{act}}\ge 2$, draw $M$ i.i.d. unit vectors $\{\mathbf u_m^{(c)}\}_{m=1}^{M}$ uniformly on $\mathbb S^{r_{\mathrm{act}}-1}$. Let $p_{r_{\mathrm{act}}}(\theta)=\frac{1}{2}I_{\sin^2\theta}\!\left(\frac{r_{\mathrm{act}}-1}{2},\frac{1}{2}\right)$ be the spherical-cap probability of falling within angle $\theta\in(0,\pi/2]$ of a fixed direction, where $I_x(\cdot,\cdot)$ is the regularized incomplete beta function. Then $\Pr\!\left(\min_{1\le m\le M}\angle(\mathbf u_m^{(c)},\mathbf u_1)\le \theta\right)
=1-\left(1-p_{r_{\mathrm{act}}}(\theta)\right)^M$. Therefore, for $r_{\mathrm{act}}=1$, one candidate suffices.
\end{theorem}

\begin{lemma}[Directional Suboptimality]\label{lem:subopt}
Let $\mathbf H_T\succeq 0$ have largest eigenvalue $\lambda_1$ and top eigenvector $\mathbf u_1$. For any unit $\widehat{\mathbf u}$ with $\angle(\widehat{\mathbf u},\mathbf u_1)\le\theta$,
\[
\widehat{\mathbf u}^{\top}\mathbf H_T\widehat{\mathbf u}\ge \lambda_1\cos^2\theta,
\qquad
\lambda_1-\widehat{\mathbf u}^{\top}\mathbf H_T\widehat{\mathbf u}\le \lambda_1\sin^2\theta .
\]
\end{lemma}

\begin{corollary}[Candidate Budget]\label{col:M}
For $r_{\mathrm{act}}\ge 2$, to obtain coverage probability at least $1-\delta$ within angle $\theta$, it suffices to choose $M\ge
\left\lceil
\frac{\log \delta}{\log\left(1-p_{r_{\mathrm{act}}}(\theta)\right)}
\right\rceil$. Together with Lemma~\ref{lem:subopt}, this gives an explicit tradeoff between candidate count and local directional quality.
\end{corollary}

\section{Empirical Studies of the Effectiveness of SOMA}\label{sec:exp}

We evaluate SOMA through five questions. \textbf{RQ1}: Can a small surrogate preserve the original model's responses? \textbf{RQ2}: Does higher response similarity also improve task-grounded quality? \textbf{RQ3}: When does SOMA provide net efficiency gains after warm-start and adaptation overhead? \textbf{RQ4}: Which components of SOMA are most important? \textbf{RQ5}: When is session-local serving reliable, and when should SOMA fall back to the large model?

\textbf{Datasets.} We evaluate SOMA on six multi-turn benchmarks: ShareGPT~\citep{chen2024sharegpt4v}, ReMeDi~\citep{yan2022remedi}, Craigslist~\citep{he2018decoupling}, Multi-Char~\citep{agentlans_multi-character_dialogue_2024}, MATH~\citep{hendrycks2021measuring}, and MT-Bench~\citep{zheng2023judging}. Dataset details are in Appendix~\ref{sec:app_dataset}.

\textbf{Models and baselines.} We test two model families: LLaMA, using LLaMA-3.1-70B as \(F\) and LLaMA-2-7B as \(G\), and Qwen, using Qwen-3-8B as \(F\) and Qwen-3-0.6B as \(G\). We compare with Original, Surrogate, History-Prefix, History-FT, LLMLingua-2~\citep{pan2024llmlingua}, and RouteLLM~\citep{ong2024routellmlearningroutellms}. We also include Random-FT, which uses the same LoRA budget as SOMA but samples local training turns uniformly, to isolate the effect of soft-prompt mining.

\textbf{Metrics.} We mainly measure teacher-fidelity similarity, i.e., how closely each method preserves the original model's response, using three LLM judges: GPT-OSS~\citep{openai2025gptoss120bgptoss20bmodel}, DeepSeek-V3, and Gemma-2-27B. To complement similarity with task-grounded quality, we report EM accuracy on MATH. Efficiency is measured by token usage, throughput, and end-to-end latency, including warm-start and adaptation overhead.

\textbf{Implementation.} We set the warm-start window, acceptance batch, prompt candidates, and switch threshold according to the analysis in Section~\ref{sec:theory}. After switching, SOMA discards soft prompts, performs no further LoRA updates, and only keeps the lightweight drift check in Section~\ref{sec:method}. More details are in Appendix~\ref{sec:app_implementation}.

\subsection{Experimental Results}
\label{sec:exp_res}

\paragraph{\textbf{RQ1: Response similarity to the original model.}}
SOMA is designed as a drop-in replacement for large-model serving. After a warm-start stage, the small surrogate should preserve the original model's behavior within the current dialogue state. Table~\ref{tab:llama-pct} reports response similarity on the LLaMA family. SOMA achieves the highest average similarity across all six datasets. The gap over Surrogate shows the behavior loss caused by directly replacing the large model with a smaller model. The gap over History-Prefix shows that simply giving the small model more history is insufficient. The gap over History-FT shows that local adaptation helps, but becomes stronger when SOMA selects training examples from weak-alignment directions. The largest gains appear on MATH and Multi-Char, where later turns depend on accumulated reasoning states, constraints, and roles. Full Qwen-family results show the same trend and are reported in Appendix~\ref{sec:app_res}.

\begin{table}[t]
\centering
\caption{Response similarity to the original model on the LLaMA family. Higher is better.}
\label{tab:llama-pct}
\resizebox{\textwidth}{!}{
\begin{tabular}{lccccccc}
\toprule
 & \textbf{ShareGPT} & \textbf{ReMeDi} & \textbf{Craigslist} & \textbf{Multi-Char} & \textbf{MATH} & \textbf{MT-Bench} & \textbf{Avg.} \\
\midrule
Surrogate      & 79.2 $\pm$ 2.18 & 82.7 $\pm$ 1.95 & 74.3 $\pm$ 2.36 & 70.8 $\pm$ 1.73 & 66.2 $\pm$ 2.91 & 77.5 $\pm$ 2.04 & 75.1 $\pm$ 5.98 \\
History-Prefix & 86.1 $\pm$ 1.67 & 87.9 $\pm$ 1.55 & 82.4 $\pm$ 1.72 & 84.7 $\pm$ 1.69 & 80.3 $\pm$ 3.83 & 87.2 $\pm$ 1.58 & 84.8 $\pm$ 2.94 \\
History-FT     & 93.4 $\pm$ 2.12 & 91.8 $\pm$ 2.09 & 90.3 $\pm$ 1.23 & 89.1 $\pm$ 2.23 & 87.6 $\pm$ 2.26 & 92.4 $\pm$ 1.14 & 90.8 $\pm$ 2.18 \\
LLMLingua-2    & 84.6 $\pm$ 1.72 & 86.4 $\pm$ 1.63 & 80.8 $\pm$ 1.91 & 82.9 $\pm$ 1.58 & 78.1 $\pm$ 2.77 & 85.3 $\pm$ 1.66 & 83.0 $\pm$ 3.03 \\
RouteLLM       & 95.3 $\pm$ 1.44 & 92.5 $\pm$ 1.07 & 91.0 $\pm$ 1.86 & 91.4 $\pm$ 1.23 & 89.6 $\pm$ 1.95 & 93.2 $\pm$ 1.12 & 92.2 $\pm$ 1.78 \\
\midrule
\textbf{SOMA}  & \textbf{96.4 $\pm$ 1.91} & \textbf{93.2 $\pm$ 0.98} & \textbf{91.9 $\pm$ 2.49} & \textbf{92.3 $\pm$ 1.05} & \textbf{90.7 $\pm$ 1.12} & \textbf{94.1 $\pm$ 0.91} & \textbf{93.1 $\pm$ 1.99} \\
\bottomrule
\end{tabular}}
\end{table}

\paragraph{\textbf{RQ2: Task-grounded quality.}}
Response similarity is the primary metric for replacing the original model, but similarity alone does not guarantee task quality. We therefore evaluate MATH using exact-match accuracy. Table~\ref{tab:math-em} shows that SOMA remains below the original large model, as expected, but substantially narrows the gap between the surrogate and the original model. SOMA also outperforms History-Prefix, History-FT, and RouteLLM in both model families. This suggests that SOMA's similarity gains are not only surface-level matching; local adaptation also preserves more of the original model's reasoning behavior.

\begin{table}[t]
\centering
\caption{MATH exact-match accuracy. SOMA improves task-grounded quality over baselines.}
\label{tab:math-em}
\resizebox{0.78\textwidth}{!}{
\begin{tabular}{lcccccc}
\toprule
\textbf{Family} & \textbf{Orig.} & \textbf{Surr.} & \textbf{Hist.-P} & \textbf{Hist.-FT} & \textbf{Route} & \textbf{SOMA} \\
\midrule
LLaMA & 48.34 $\pm$ 0.32 & 19.20 $\pm$ 0.78 & 25.03 $\pm$ 0.94 & 31.46 $\pm$ 0.87 & 33.88 $\pm$ 0.79 & \textbf{41.62 $\pm$ 0.66} \\
Qwen  & 36.48 $\pm$ 0.41 & 11.73 $\pm$ 0.64 & 16.21 $\pm$ 0.79 & 22.57 $\pm$ 0.83 & 25.08 $\pm$ 0.71 & \textbf{31.14 $\pm$ 0.74} \\
\bottomrule
\end{tabular}}
\end{table}

\paragraph{\textbf{RQ3: End-to-end efficiency.}}
\begin{wraptable}[9]{r}{0.36\textwidth}
\vspace{-2em}
\centering
\caption{ShareGPT break-even after all overhead.}
\label{tab:break-even}
\vspace{3pt}
\setlength{\tabcolsep}{1pt}
\begin{tabular}{lcc}
\toprule
\textbf{Turns} & \textbf{Latency} & \textbf{Tokens} \\
\midrule
1--4  & $-4.1\%$  & $-1.2\%$ \\
5--8  & $+4.7\%$  & $+8.5\%$ \\
9--12 & $+16.3\%$ & $+23.8\%$ \\
13+   & $+29.1\%$ & $+37.2\%$ \\
\bottomrule
\end{tabular}
\vspace{-10pt}
\end{wraptable}

\begin{figure}[t]
  \centering
  \begin{subfigure}[t]{0.42\textwidth}
    \centering
    \includegraphics[width=\linewidth]{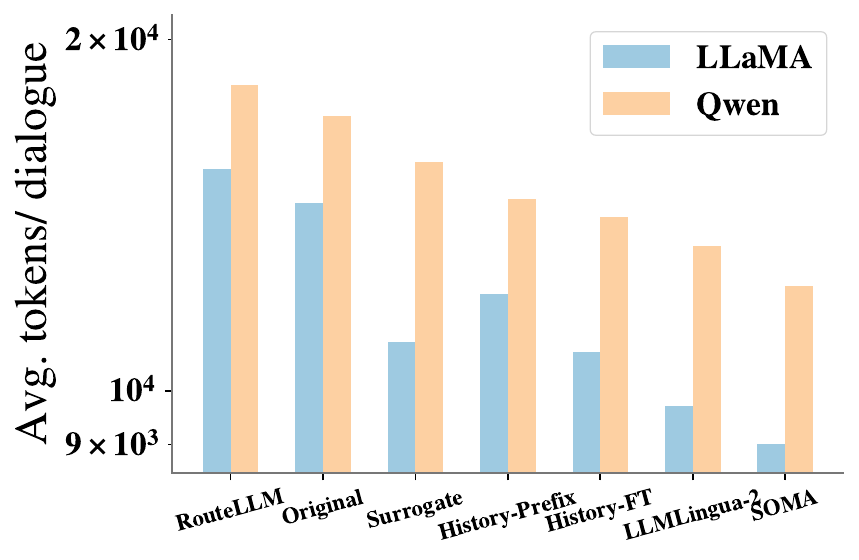}
    \caption{Token usage.}
    \label{fig:token_usage}
  \end{subfigure}\hfill
  \begin{subfigure}[t]{0.54\textwidth}
    \centering
    \includegraphics[width=\linewidth]{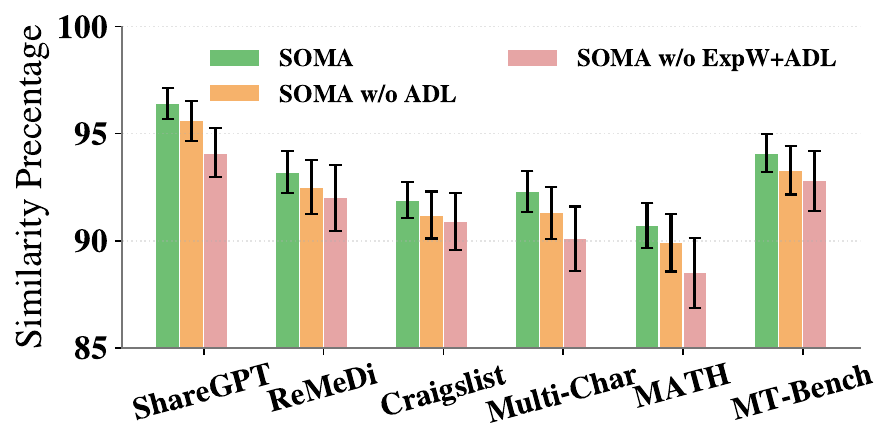}
    \caption{LLaMA ablation.}
    \label{fig:llama_ablation}
  \end{subfigure}
  \vspace{-2mm}
  \caption{Efficiency and component analysis. SOMA reduces token usage after switching, and the ablation shows that both anti-degeneration regularization and expectation-weighted divergence improve soft-prompt mining.}
  \label{fig:token_ablation}
\end{figure}

SOMA reduces post-switch cost by serving the adapted surrogate with compressed context, but it also introduces one-time costs from soft-prompt probing and LoRA adaptation. Thus, the key question is whether the total session cost decreases after this overhead is included. Figure~\ref{fig:token_usage} shows that SOMA reduces average token usage by avoiding repeated full-history serving after switching. Table~\ref{tab:break-even} further shows the amortized operating range: SOMA is slightly unfavorable for very short sessions, starts to help on medium-length sessions, and becomes clearly beneficial for longer sessions. This matches SOMA's intended use case as an amortized serving strategy for conversations where enough later turns remain local to the warm-start state.

\paragraph{\textbf{RQ4: Which components of SOMA are most important?}}
We first ablate SOMA on the LLaMA family to identify which components drive the gains. Figure~\ref{fig:llama_ablation} shows that full SOMA performs best across all datasets. Removing the anti-degeneration loss consistently reduces performance, which shows that entropy regularization is important for keeping prompt mining stable and preventing collapsed or low-diversity soft prompts. Removing both the expectation-weighted term and the anti-degeneration loss leads to a larger drop, especially on harder datasets. This result shows that SOMA benefits from both stable prompt search and disagreement-aware mining. In other words, SOMA works not only because it adapts the surrogate locally, but also because it mines more informative weak-alignment directions before adaptation.


\paragraph{\textbf{RQ5: When is session-local serving reliable, and when should SOMA fall back?}}
SOMA is intended for dialogue stretches that remain tied to the warm-start state, so we next test when this assumption holds and how rollback helps when it fails. Table~\ref{tab:context-dependency} shows that full-history MATH performance stays stable across turn buckets, while no-history performance drops sharply in later turns. This means later turns are not simply easier; they remain strongly dependent on the earlier dialogue context. This is exactly the setting SOMA is designed for. Table~\ref{tab:topic-shift} then evaluates abrupt topic shifts. Without rollback, similarity drops substantially. With SOMA's drift-aware rollback, most of the loss is recovered while false rollback remains low. These results show that SOMA should be viewed as a local serving strategy with automatic fallback.

\begin{table}[t]
\centering
\caption{Reliability analysis for session-local serving. Later turns remain strongly context-dependent, and drift-aware rollback recovers most of the loss under abrupt topic shifts.}
\begin{subtable}[t]{0.43\textwidth}
\centering
\caption{MATH context dependency.}
\label{tab:context-dependency}
\resizebox{\linewidth}{!}{
\begin{tabular}{lccc}
\toprule
\textbf{Setting} & \textbf{1--3} & \textbf{4--6} & \textbf{7+} \\
\midrule
Full history & 46.92 $\pm$ 0.58 & 47.18 $\pm$ 0.51 & 46.75 $\pm$ 0.49 \\
No history   & 43.78 $\pm$ 0.66 & 34.92 $\pm$ 0.76 & 24.88 $\pm$ 0.86 \\
Drop         & 3.14 & 12.26 & 21.87 \\
\bottomrule
\end{tabular}}
\end{subtable}
\hfill
\begin{subtable}[t]{0.54\textwidth}
\centering
\caption{Topic-shift stress test.}
\label{tab:topic-shift}
\resizebox{\linewidth}{!}{
\begin{tabular}{lccccc}
\toprule
\textbf{Data} & \textbf{Clean} & \textbf{No RB} & \textbf{SOMA} & \textbf{Detect} & \textbf{False RB} \\
\midrule
ShareGPT   & 96.4 $\pm$ 1.91 & 88.5 $\pm$ 2.03 & \textbf{94.1 $\pm$ 1.47} & 90.8\% & 4.3\% \\
ReMeDi     & 93.2 $\pm$ 0.98 & 85.7 $\pm$ 1.92 & \textbf{91.1 $\pm$ 1.34} & 89.6\% & 3.9\% \\
Craigslist & 91.9 $\pm$ 2.49 & 83.1 $\pm$ 2.11 & \textbf{89.3 $\pm$ 1.68} & 88.9\% & 3.8\% \\
\bottomrule
\end{tabular}}
\end{subtable}
\vspace{-2mm}
\label{tab:reliability}
\end{table}
\vspace{-0.2em}
\section{Conclusion}\label{sec:conclusion}
\vspace{-0.2em}
We presented SOMA, an efficient multi-turn LLM serving framework that uses early dialogue turns to adapt a small surrogate to the original model's session-local behavior. SOMA mines informative soft-prompt directions, distills them with lightweight LoRA adaptation, and switches to the surrogate only when semantic alignment is sufficient. Experiments show that SOMA better preserves original-model responses, improves task-grounded quality, and reduces serving cost once warm-start overhead is amortized. These results suggest that session-local surrogate serving is a practical direction for efficient multi-turn LLM deployment.

\bibliographystyle{plain}
\bibliography{ref}

\begin{thebibliography}{10}

\bibitem{achiam2023gpt}
Josh Achiam, Steven Adler, Sandhini Agarwal, Lama Ahmad, Ilge Akkaya, Florencia~Leoni Aleman, Diogo Almeida, Janko Altenschmidt, Sam Altman, Shyamal Anadkat, et~al.
\newblock Gpt-4 technical report.
\newblock {\em arXiv preprint arXiv:2303.08774}, 2023.

\bibitem{agentlans_multi-character_dialogue_2024}
agentlans.
\newblock Multi-character dialogue dataset.
\newblock Hugging Face Dataset, 2024.
\newblock CC-BY-4.0 license.

\bibitem{ananda2025beyond}
Shalini Ananda.
\newblock Beyond the bubble: How context‑aware memory systems are changing the game in 2025.
\newblock Tribe AI Applied AI Blog, April 2025.

\bibitem{anthropic_introducing_claude_2023}
Anthropic.
\newblock Introducing claude.
\newblock Anthropic Blog, March 2023.
\newblock \url{https://www.anthropic.com/news/introducing-claude}.

\bibitem{anthropic_prompt_caching_2024}
{Anthropic}.
\newblock Prompt caching with claude.
\newblock \url{https://www.anthropic.com/news/prompt-caching}, August 2024.
\newblock Accessed 2025-06-12.

\bibitem{bai2024mt}
Ge~Bai, Jie Liu, Xingyuan Bu, Yancheng He, Jiaheng Liu, Zhanhui Zhou, Zhuoran Lin, Wenbo Su, Tiezheng Ge, Bo~Zheng, et~al.
\newblock Mt-bench-101: A fine-grained benchmark for evaluating large language models in multi-turn dialogues.
\newblock {\em arXiv preprint arXiv:2402.14762}, 2024.

\bibitem{behera2025towards}
Adarsh~Prasad Behera, Jaya~Prakash Champati, Roberto Morabito, Sasu Tarkoma, and James Gross.
\newblock Towards efficient multi-llm inference: Characterization and analysis of llm routing and hierarchical techniques.
\newblock {\em arXiv preprint arXiv:2506.06579}, 2025.

\bibitem{belkin2003laplacian}
Mikhail Belkin and Partha Niyogi.
\newblock Laplacian eigenmaps for dimensionality reduction and data representation.
\newblock {\em Neural computation}, 15(6):1373--1396, 2003.

\bibitem{brown2020language}
Tom Brown, Benjamin Mann, Nick Ryder, Melanie Subbiah, Jared~D Kaplan, Prafulla Dhariwal, Arvind Neelakantan, Pranav Shyam, Girish Sastry, Amanda Askell, et~al.
\newblock Language models are few-shot learners.
\newblock {\em Advances in neural information processing systems}, 33:1877--1901, 2020.

\bibitem{chen2024sharegpt4v}
Lin Chen, Jinsong Li, Xiaoyi Dong, Pan Zhang, Conghui He, Jiaqi Wang, Feng Zhao, and Dahua Lin.
\newblock Sharegpt4v: Improving large multi-modal models with better captions.
\newblock In {\em European Conference on Computer Vision}, pages 370--387. Springer, 2024.

\bibitem{chen2024compress}
Nuo Chen, Hongguang Li, Juhua Huang, Baoyuan Wang, and Jia Li.
\newblock Compress to impress: Unleashing the potential of compressive memory in real-world long-term conversations.
\newblock {\em arXiv preprint arXiv:2402.11975}, 2024.

\bibitem{coifman2006diffusion}
Ronald~R Coifman and St{\'e}phane Lafon.
\newblock Diffusion maps.
\newblock {\em Applied and computational harmonic analysis}, 21(1):5--30, 2006.

\bibitem{ding2024hybrid}
Dujian Ding, Ankur Mallick, Chi Wang, Robert Sim, Subhabrata Mukherjee, Victor Ruhle, Laks~VS Lakshmanan, and Ahmed~Hassan Awadallah.
\newblock Hybrid llm: Cost-efficient and quality-aware query routing.
\newblock {\em arXiv preprint arXiv:2404.14618}, 2024.

\bibitem{dong2023towards}
Xin~Luna Dong, Seungwhan Moon, Yifan~Ethan Xu, Kshitiz Malik, and Zhou Yu.
\newblock Towards next-generation intelligent assistants leveraging llm techniques.
\newblock In {\em Proceedings of the 29th ACM SIGKDD Conference on Knowledge Discovery and Data Mining}, pages 5792--5793, 2023.

\bibitem{fang2025non}
Yao~Hui Fang and Xing~Ce Wang.
\newblock Non-uniform point cloud upsampling via local manifold distribution.
\newblock {\em Proceedings of the ACM on Computer Graphics and Interactive Techniques}, 8(1):1--15, 2025.

\bibitem{gao2024cost}
Bin Gao, Zhuomin He, Puru Sharma, Qingxuan Kang, Djordje Jevdjic, Junbo Deng, Xingkun Yang, Zhou Yu, and Pengfei Zuo.
\newblock $\{$Cost-Efficient$\}$ large language model serving for multi-turn conversations with $\{$CachedAttention$\}$.
\newblock In {\em 2024 USENIX Annual Technical Conference (USENIX ATC 24)}, pages 111--126, 2024.

\bibitem{guo2025deepseek}
Daya Guo, Dejian Yang, Haowei Zhang, Junxiao Song, Ruoyu Zhang, Runxin Xu, Qihao Zhu, Shirong Ma, Peiyi Wang, Xiao Bi, et~al.
\newblock Deepseek-r1: Incentivizing reasoning capability in llms via reinforcement learning.
\newblock {\em arXiv preprint arXiv:2501.12948}, 2025.

\bibitem{gutierrez2024hipporag}
Bernal~Jim{\'e}nez Guti{\'e}rrez, Yiheng Shu, Yu~Gu, Michihiro Yasunaga, and Yu~Su.
\newblock Hipporag: Neurobiologically inspired long-term memory for large language models.
\newblock In {\em The Thirty-eighth Annual Conference on Neural Information Processing Systems}, 2024.

\bibitem{he2018decoupling}
He~He, Derek Chen, Anusha Balakrishnan, and Percy Liang.
\newblock Decoupling strategy and generation in negotiation dialogues.
\newblock {\em arXiv preprint arXiv:1808.09637}, 2018.

\bibitem{hendrycks2021measuring}
Dan Hendrycks, Collin Burns, Saurav Kadavath, Akul Arora, Steven Basart, Eric Tang, Dawn Song, and Jacob Steinhardt.
\newblock Measuring mathematical problem solving with the math dataset.
\newblock {\em arXiv preprint arXiv:2103.03874}, 2021.

\bibitem{hinton2015distilling}
Geoffrey Hinton, Oriol Vinyals, and Jeff Dean.
\newblock Distilling the knowledge in a neural network.
\newblock {\em arXiv preprint arXiv:1503.02531}, 2015.

\bibitem{holtzman2019curious}
Ari Holtzman, Jan Buys, Li~Du, Maxwell Forbes, and Yejin Choi.
\newblock The curious case of neural text degeneration.
\newblock {\em arXiv preprint arXiv:1904.09751}, 2019.

\bibitem{hu2022lora}
Edward~J Hu, Yelong Shen, Phillip Wallis, Zeyuan Allen-Zhu, Yuanzhi Li, Shean Wang, Lu~Wang, Weizhu Chen, et~al.
\newblock Lora: Low-rank adaptation of large language models.
\newblock {\em ICLR}, 1(2):3, 2022.

\bibitem{jeong2025accelerating}
Jinwoo Jeong and Jeongseob Ahn.
\newblock Accelerating llm serving for multi-turn dialogues with efficient resource management.
\newblock In {\em Proceedings of the 30th ACM International Conference on Architectural Support for Programming Languages and Operating Systems, Volume 2}, pages 1--15, 2025.

\bibitem{laban2025llms}
Philippe Laban, Hiroaki Hayashi, Yingbo Zhou, and Jennifer Neville.
\newblock Llms get lost in multi-turn conversation.
\newblock {\em arXiv preprint arXiv:2505.06120}, 2025.

\bibitem{li2023repetition}
Huayang Li, Tian Lan, Zihao Fu, Deng Cai, Lemao Liu, Nigel Collier, Taro Watanabe, and Yixuan Su.
\newblock Repetition in repetition out: Towards understanding neural text degeneration from the data perspective.
\newblock {\em Advances in Neural Information Processing Systems}, 36:72888--72903, 2023.

\bibitem{li2026llmclinicalgraphstructure}
Lincan Li, Zheng Chen, and Yushun Dong.
\newblock Llm as clinical graph structure refiner: Enhancing representation learning in eeg seizure diagnosis, 2026.

\bibitem{li2025beyond}
Yubo Li, Xiaobin Shen, Xinyu Yao, Xueying Ding, Yidi Miao, Ramayya Krishnan, and Rema Padman.
\newblock Beyond single-turn: A survey on multi-turn interactions with large language models.
\newblock {\em arXiv preprint arXiv:2504.04717}, 2025.

\bibitem{liu2024exploring}
Fang Liu, Yang Liu, Lin Shi, Houkun Huang, Ruifeng Wang, Zhen Yang, Li~Zhang, Zhongqi Li, and Yuchi Ma.
\newblock Exploring and evaluating hallucinations in llm-powered code generation.
\newblock {\em arXiv preprint arXiv:2404.00971}, 2024.

\bibitem{mcinnes2018umap}
Leland McInnes, John Healy, and James Melville.
\newblock Umap: Uniform manifold approximation and projection for dimension reduction.
\newblock {\em arXiv preprint arXiv:1802.03426}, 2018.

\bibitem{meister2023locally}
Clara Meister, Tiago Pimentel, Gian Wiher, and Ryan Cotterell.
\newblock Locally typical sampling.
\newblock {\em Transactions of the Association for Computational Linguistics}, 11:102--121, 2023.

\bibitem{melz2023enhancing}
Eric Melz.
\newblock Enhancing llm intelligence with arm-rag: Auxiliary rationale memory for retrieval augmented generation.
\newblock {\em arXiv preprint arXiv:2311.04177}, 2023.

\bibitem{moon2025accelerating}
Don Moon.
\newblock Accelerating multi‑turn llm serving with multi‑tier caching and smarter scheduling.
\newblock Medium (Byte-Sized AI), April 2025.

\bibitem{ngo2023enhancing}
Giang Ngo and Nhi~NY Vo.
\newblock Enhancing recommendation systems with hybrid manifold regularized knowledge graph.
\newblock In {\em 2023 IEEE 10th International Conference on Data Science and Advanced Analytics (DSAA)}, pages 1--8. IEEE, 2023.

\bibitem{ong2024routellmlearningroutellms}
Isaac Ong, Amjad Almahairi, Vincent Wu, Wei-Lin Chiang, Tianhao Wu, Joseph~E. Gonzalez, M~Waleed Kadous, and Ion Stoica.
\newblock Routellm: Learning to route llms with preference data, 2024.

\bibitem{openai2025gptoss120bgptoss20bmodel}
OpenAI.
\newblock gpt-oss-120b \& gpt-oss-20b model card, 2025.

\bibitem{pan2024llmlingua}
Zhuoshi Pan, Qianhui Wu, Huiqiang Jiang, Menglin Xia, Xufang Luo, Jue Zhang, Qingwei Lin, Victor R{\"u}hle, Yuqing Yang, Chin-Yew Lin, et~al.
\newblock Llmlingua-2: Data distillation for efficient and faithful task-agnostic prompt compression.
\newblock {\em arXiv preprint arXiv:2403.12968}, 2024.

\bibitem{park2023thinking}
Soya Park and Chinmay Kulkarni.
\newblock Thinking assistants: Llm-based conversational assistants that help users think by asking rather than answering.
\newblock {\em arXiv preprint arXiv:2312.06024}, 2023.

\bibitem{radford2019language}
Alec Radford, Jeffrey Wu, Rewon Child, David Luan, Dario Amodei, Ilya Sutskever, et~al.
\newblock Language models are unsupervised multitask learners.
\newblock {\em OpenAI blog}, 1(8):9, 2019.

\bibitem{roweis2000nonlinear}
Sam~T Roweis and Lawrence~K Saul.
\newblock Nonlinear dimensionality reduction by locally linear embedding.
\newblock {\em science}, 290(5500):2323--2326, 2000.

\bibitem{schick2023toolformer}
Timo Schick, Jane Dwivedi-Yu, Roberto Dess{\`\i}, Roberta Raileanu, Maria Lomeli, Eric Hambro, Luke Zettlemoyer, Nicola Cancedda, and Thomas Scialom.
\newblock Toolformer: Language models can teach themselves to use tools.
\newblock {\em Advances in Neural Information Processing Systems}, 36:68539--68551, 2023.

\bibitem{shah2024flashattention}
Jay Shah, Ganesh Bikshandi, Ying Zhang, Vijay Thakkar, Pradeep Ramani, and Tri Dao.
\newblock Flashattention-3: Fast and accurate attention with asynchrony and low-precision.
\newblock {\em Advances in Neural Information Processing Systems}, 37:68658--68685, 2024.

\bibitem{shnitzer2023large}
Tal Shnitzer, Anthony Ou, M{\'\i}rian Silva, Kate Soule, Yuekai Sun, Justin Solomon, Neil Thompson, and Mikhail Yurochkin.
\newblock Large language model routing with benchmark datasets.
\newblock {\em arXiv preprint arXiv:2309.15789}, 2023.

\bibitem{stolcke2000dialogue}
Andreas Stolcke, Klaus Ries, Noah Coccaro, Elizabeth Shriberg, Rebecca Bates, Daniel Jurafsky, Paul Taylor, Rachel Martin, Carol~Van Ess-Dykema, and Marie Meteer.
\newblock Dialogue act modeling for automatic tagging and recognition of conversational speech.
\newblock {\em Computational linguistics}, 26(3):339--373, 2000.

\bibitem{sun2020zernet}
Zhiyu Sun, Ethan Rooke, Jerome Charton, Yusen He, Jia Lu, and Stephen Baek.
\newblock Zernet: Convolutional neural networks on arbitrary surfaces via zernike local tangent space estimation.
\newblock In {\em Computer Graphics Forum}, volume~39, pages 204--216. Wiley Online Library, 2020.

\bibitem{touvron2023llama}
Hugo Touvron, Thibaut Lavril, Gautier Izacard, Xavier Martinet, Marie-Anne Lachaux, Timoth{\'e}e Lacroix, Baptiste Rozi{\`e}re, Naman Goyal, Eric Hambro, Faisal Azhar, et~al.
\newblock Llama: Open and efficient foundation language models.
\newblock {\em arXiv preprint arXiv:2302.13971}, 2023.

\bibitem{van2008visualizing}
Laurens Van~der Maaten and Geoffrey Hinton.
\newblock Visualizing data using t-sne.
\newblock {\em Journal of machine learning research}, 9(11), 2008.

\bibitem{wang2025recursively}
Qingyue Wang, Yanhe Fu, Yanan Cao, Shuai Wang, Zhiliang Tian, and Liang Ding.
\newblock Recursively summarizing enables long-term dialogue memory in large language models.
\newblock {\em Neurocomputing}, 639:130193, 2025.

\bibitem{welleck2019neural}
Sean Welleck, Ilia Kulikov, Stephen Roller, Emily Dinan, Kyunghyun Cho, and Jason Weston.
\newblock Neural text generation with unlikelihood training.
\newblock {\em arXiv preprint arXiv:1908.04319}, 2019.

\bibitem{xiao2024efficient}
Guangxuan Xiao, Yuandong Tian, Beidi Chen, Song Han, and Mike Lewis.
\newblock Efficient streaming language models with attention sinks, 2024.
\newblock {\em URL https://arxiv. org/abs/2309.17453}, page~1, 2024.

\bibitem{xiong2020loco}
Yuwen Xiong, Mengye Ren, and Raquel Urtasun.
\newblock Loco: Local contrastive representation learning.
\newblock {\em Advances in neural information processing systems}, 33:11142--11153, 2020.

\bibitem{yan2022remedi}
Guojun Yan, Jiahuan Pei, Pengjie Ren, Zhaochun Ren, Xin Xin, Huasheng Liang, Maarten De~Rijke, and Zhumin Chen.
\newblock Remedi: Resources for multi-domain, multi-service, medical dialogues.
\newblock In {\em Proceedings of the 45th International ACM SIGIR Conference on Research and Development in Information Retrieval}, pages 3013--3024, 2022.

\bibitem{yi2024survey}
Zihao Yi, Jiarui Ouyang, Yuwen Liu, Tianhao Liao, Zhe Xu, and Ying Shen.
\newblock A survey on recent advances in llm-based multi-turn dialogue systems.
\newblock {\em arXiv preprint arXiv:2402.18013}, 2024.

\bibitem{yu2026health}
Dahai Yu, Lin Jiang, Rongchao Xu, and Guang Wang.
\newblock Healthmamba: An uncertainty-aware spatiotemporal graph state space model for effective and reliable healthcare facility visit prediction, 2026.

\bibitem{zeng2021contrastive}
Zhaoyang Zeng, Daniel McDuff, Yale Song, et~al.
\newblock Contrastive learning of global and local video representations.
\newblock {\em Advances in Neural Information Processing Systems}, 34:7025--7040, 2021.

\bibitem{zhang2007linear}
Tianhao Zhang, Jie Yang, Deli Zhao, and Xinliang Ge.
\newblock Linear local tangent space alignment and application to face recognition.
\newblock {\em Neurocomputing}, 70(7-9):1547--1553, 2007.

\bibitem{zhao2025amplifying}
Yuying Zhao, Yu~Wang, Xueqi Cheng, Anne~Marie Tumlin, Yunchao Liu, Damin Xia, Meng Jiang, and Tyler Derr.
\newblock Amplifying your social media presence: Personalized influential content generation with llms.
\newblock {\em arXiv preprint arXiv:2505.01698}, 2025.

\bibitem{zheng2023judging}
Lianmin Zheng, Wei-Lin Chiang, Ying Sheng, Siyuan Zhuang, Zhanghao Wu, Yonghao Zhuang, Zi~Lin, Zhuohan Li, Dacheng Li, Eric Xing, et~al.
\newblock Judging llm-as-a-judge with mt-bench and chatbot arena.
\newblock {\em Advances in neural information processing systems}, 36:46595--46623, 2023.

\end{thebibliography}

\appendix

\newpage
\section*{Technical Appendices and Supplementary Material}

\section{Notations}\label{sec:notation}

This section summarizes all notations used throughout this paper.

\begin{table}[h]
\centering
\caption{Notation summary.}
\label{tab:notation}
\setlength{\tabcolsep}{6pt}
\begin{tabular}{@{}ll@{}}
\toprule
Symbol & Meaning \\
\midrule
$\mathcal{D}_k$ & Dialogue prefix $\{(q_t,a_t)\}_{t=1}^k$ \\
$q_{\le t}$ & Full history up to turn $t$ \\
$F,\,G$ & Original (large) model; surrogate (small) model \\
$\mathbf{h}_t=f_M(q_{\le t})$ & Hidden state at turn $t$ for model $M$ \\
$\mathcal{H}_k$ & $\{\mathbf{h}_1,\ldots,\mathbf{h}_k\}$, first-$k$ hidden states \\
$\mathcal{M}_k^M$ & Local response manifold induced by $\mathcal{D}_k$ under $M$ \\
$\mathcal{V}$, $E$ & $G$’s vocabulary; embedding matrix $E\in\mathbb{R}^{d\times|\mathcal{V}|}$ \\
$\Pi_t$ & Next-token distribution of $G$ at turn $t$; entry $\Pi_t[v]$ \\
$\mathbf{e}_v$ & Embedding of token $v$ (column of $E$) \\
$\mathbf{P}\in\mathbb{R}^{L\times d}$ & Soft prompt (row-wise prompt tokens in embedding space) \\
$\mathsf{V}(\mathbf{P})$ & Verbalization for prompt $\mathbf{P}$\\
$N_k(u)$ & $k$ nearest neighbors of token $u$ by cosine in $E$ \\
$\overline{\mathbf{e}}_{t,i}$ & Expected embedding $\overline{\mathbf{e}}_{t,i}=E^\top \Pi_{t,i}$ at position $i$ \\
$w_{t,i}$ & Expectation-weighted factor for semantic divergence loss \\
$|B|_{\mathrm{eff}}$ & Effective batch size (accounts for dependence) \\
$\Delta$ & Empirical improvement (old–new) on a post-warm-start batch \\
\bottomrule
\end{tabular}
\end{table}

\section{Related Work}\label{sec:related}

\textbf{Efficient LLM for Multi-turn Dialogues.} Recent efforts to improve multi-turn LLM serving efficiency fall into two main paradigms. The first involves \textbf{single-model approaches} that reduce context length or reuse computation. These include \textit{summarization and context compression}~\citep{wang2025recursively, chen2024compress, xiao2024efficient}, \textit{memory augmentation}~\citep{melz2023enhancing, gutierrez2024hipporag}, and \textit{caching and attention reuse}~\citep{gao2024cost, jeong2025accelerating, anthropic_prompt_caching_2024}. While effective in lowering per-turn cost, these methods still rely on repeated large-model inference, which also leads to high API expenses, latency, and GPU demand. Moreover, they may truncate or underutilize dialogue context, harming performance on complex tasks. The second paradigm adopts \textbf{multi-model approach}, using smaller models for simple queries and escalating difficult ones to larger LLMs~\citep{behera2025towards, schick2023toolformer, ding2024hybrid}, typically via model routing~\citep{shnitzer2023large} and distillation~\citep{hinton2015distilling}. However, pre-trained small models often generalize poorly on complex multi-turn dialogues, while switching models adds inefficiency. 

\textbf{Local Manifold Approximation.}  
Local manifold approximation techniques aim to exploit the manifold hypothesis by modeling high-dimensional data as lying on locally low-dimensional subspaces, enabling more efficient representation and inference. Classical approaches such as Locally Linear Embedding (LLE)~\citep{roweis2000nonlinear} and Local Tangent Space Alignment (LTSA)~\citep{zhang2007linear} approximate local neighborhoods through linear projections, while kernel-based methods like Laplacian eigenmaps~\citep{belkin2003laplacian} and diffusion maps~\citep{coifman2006diffusion} preserve local geometric structure via nonlinear embeddings. Recent work has extended these ideas using deep learning. For instance, neural network–based tangent space estimators~\citep{sun2020zernet} and local contrastive learning methods~\citep{xiong2020loco, zeng2021contrastive} enable the extraction of manifold-aware representations in complex domains. In computer vision, local manifold modeling underpins point cloud upsampling~\citep{fang2025non} by fitting Gaussian patches to local regions, while in representation learning, neighbor-preserving mappings such as t-SNE~\citep{van2008visualizing} and UMAP~\citep{mcinnes2018umap} uncover latent structure by maintaining local proximity. In graph-based learning, manifold-regularized GNNs~\citep{ngo2023enhancing} exploit smoothness over graph-induced manifolds to enhance generalization. Despite their effectiveness across domains, local manifold approximation remains largely unexplored in the context of efficient multi-turn LLM serving, where dynamically adapting smaller models to the local reasoning manifold conditioned on dialogue history presents a promising and under-investigated direction.

\section{Reproducibility}
In this section, we introduce the details of the experiments in this
paper for reproducibility. At the same time, we have uploaded all necessary code to our GitHub repository to reproduce the results presented in this paper: 
\url{https://github.com/LabRAI/SOMA}
. All major experiments are encapsulated as shell scripts, which can be conveniently executed. We introduce the details for reproducibility in the subsections below.

\subsection{Real-World Datasets}\label{sec:app_dataset}

In this section, we briefly present the real-world datasets used in this paper, and all these datasets are commonly used datasets in multi-turn conversation tasks. \textit{ShareGPT}~\citep{chen2024sharegpt4v} is a large-scale collection of high-quality image–text conversations and captions. \textit{ReMeDi}~\citep{yan2022remedi} is a multi-domain Chinese medical dialogue corpus of doctor–patient conversations. \textit{Craigslist}~\citep{he2018decoupling} contains multi-turn buyer–seller negotiation chats from Craigslist, enabling study of bargaining strategies and goal-directed dialogue. \textit{Multi-Char}~\citep{agentlans_multi-character_dialogue_2024} provides multi-character conversational scenarios with role specifications to evaluate coordination and role consistency in multi-party dialogue. \textit{MATH}~\citep{hendrycks2021measuring} is a benchmark of competition-style math problems with step-by-step solutions designed to assess mathematical reasoning in language models. \textit{MT-Bench}~\citep{zheng2023judging} is a multi-turn benchmark to assess response quality across diverse tasks. In this study, we filter out the non-context-dependent dialogues in these datasets. 

\subsection{Implementation of SOMA}\label{sec:app_implementation}

We implement SOMA based on PyTorch with HuggingFace Transformers, serve inference via vLLM with FlashAttention~\citep{shah2024flashattention}, and run on one node with $4\times$ 80G A100 GPUs. Soft–prompt tuning optimizes only the prompt tensor $\mathbf{P}\!\in\!\mathbb{R}^{L\times d}$ on the surrogate $G$ using AdamW, cosine decay, gradient clipping, and a KV cache for a single forward pass per turn. The objective combines unlikelihood on a semantic neighborhood, an expectation–weighted penalty using the truncated top–$m$ expectation, and a light anti-degeneration entropy term. The mined prompt–response pairs are then used to adapt $G$ with LoRA on attention projections (rank $r$), keeping the base weights frozen; early stopping is triggered by validation divergence. At inference, a cosine closeness gate with a lightweight sentence encoder decides a one–time switch to the adapted surrogate; the window $W$ and acceptance batch $|B|$ follow the detection bound, the number of parallel candidates $M$ follows the coverage guarantee, and $(k,m)$ follow the efficiency analysis. Prompt length $L\!\in\!\{4, 8,16,32,64\}$; learning rate $\eta\!\in\![1\!\times\!10^{-4},\,5\!\times\!10^{-3}]$; AdamW weight decay $\lambda_{\mathrm{wd}}\!\in\![0,\,10^{-2}]$; gradient clip $\in[0.5,\,1.0]$; neighborhood size $k\!\in\!\{20,50,100\}$; expectation truncation $m\!\in\!\{50,100,200\}$; temperature $\tau\!\in\![0.4,\,1.2]$; expectation weight $\lambda\!\in\![0.5,\,2.0]$; anti–degeneration weight $\beta\!\in\![0.02,\,0.15]$; LoRA rank $r\!\in\!\{4, 8,16,32\}$ with scale $\alpha_{\text{lora}}\!\in\!\{4, 8,16,32\}$; LoRA LR $\eta_{\text{lora}}\!\in\![5\!\times\!10^{-5},\,2\!\times\!10^{-3}]$; warm–start window $W\!\in\![3,\,12]$ turns; acceptance batch $|B|\!\in\![6,\,24]$ contexts; parallel candidates $M\!\in\!\{3,4,5\}$; switch threshold $\varepsilon\!\in\![0.05,\,0.12]$ with $m_{\text{cons}}\!\in\!\{2,3\}$. 

\subsection{Implementation of Baselines}\label{sec:app_baseline}

\textbf{Original}: We query the large model (family-appropriate: LLaMA-3.1-70B or Qwen-3-8B) with the full dialogue history at every turn to get the responses.

\textbf{Surrogate}: We query the small model (LLaMA-2-7B or Qwen-3-0.6B) with the full dialogue history at every turn to get the responses.

\textbf{History-Prefix}: The surrogate receives the entire conversation produced by the Original up to $t{-}1$ and generates $a_t$; no parameter updates are performed. The switching is the same as SOMA.

\textbf{History-FT}: We fine-tune the surrogate on $(S_t, a_t^{F})$ pairs where $S_t$ is the Original’s full context up to $t$ and $a_t^{F}$ is the next reply, using LoRA on attention projections with early stopping; inference then runs the fine-tuned surrogate without the Original. The switching is the same as SOMA.

\textbf{LLMLingua-2}~\citep{pan2024llmlingua}: We compress the Original’s history at each turn with LLMLingua-2 and feed the compressed summary to the surrogate; we follow the authors’ recommended settings.

\textbf{RouteLLM}~\citep{ong2024routellmlearningroutellms}: We adopt the released router to choose between the Original and the surrogate per turn (complex vs.\ simple queries), following the original settings from the paper.

\subsection{Data Filtering}

\begin{figure}[t]
    \centering
    \includegraphics[width=0.8\linewidth]{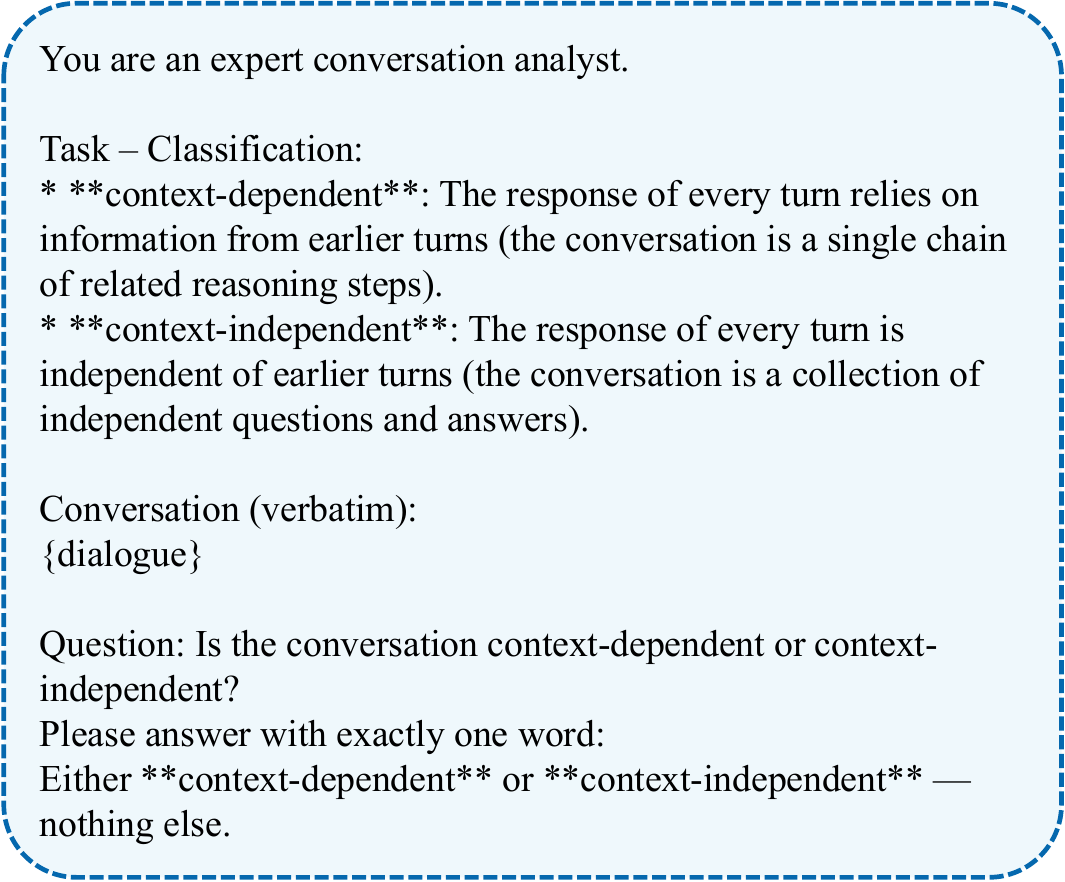}
    \caption{Instructions for the LLM to filter context-dependent dialogue.}
    \label{fig:context_filter}
\end{figure}

To ensure that SOMA is evaluated on context-dependent dialogues, we prompt multiple strong LLMs, including GPT-OSS, DeepSeek-V3, and Gemma-2-27B, with the classification prompt shown in Figure~\ref{fig:context_filter}, and retain only conversations that all three models recognize as context-dependent. This avoids model-specific biases and yields a high-quality subset of dialogues where later turns meaningfully depend on earlier turns, matching the use cases in this paper.

\subsection{Implementation of LLM Judge}\label{sec:app_judge}

\begin{figure}[t]
    \centering
    \includegraphics[width=0.8\linewidth]{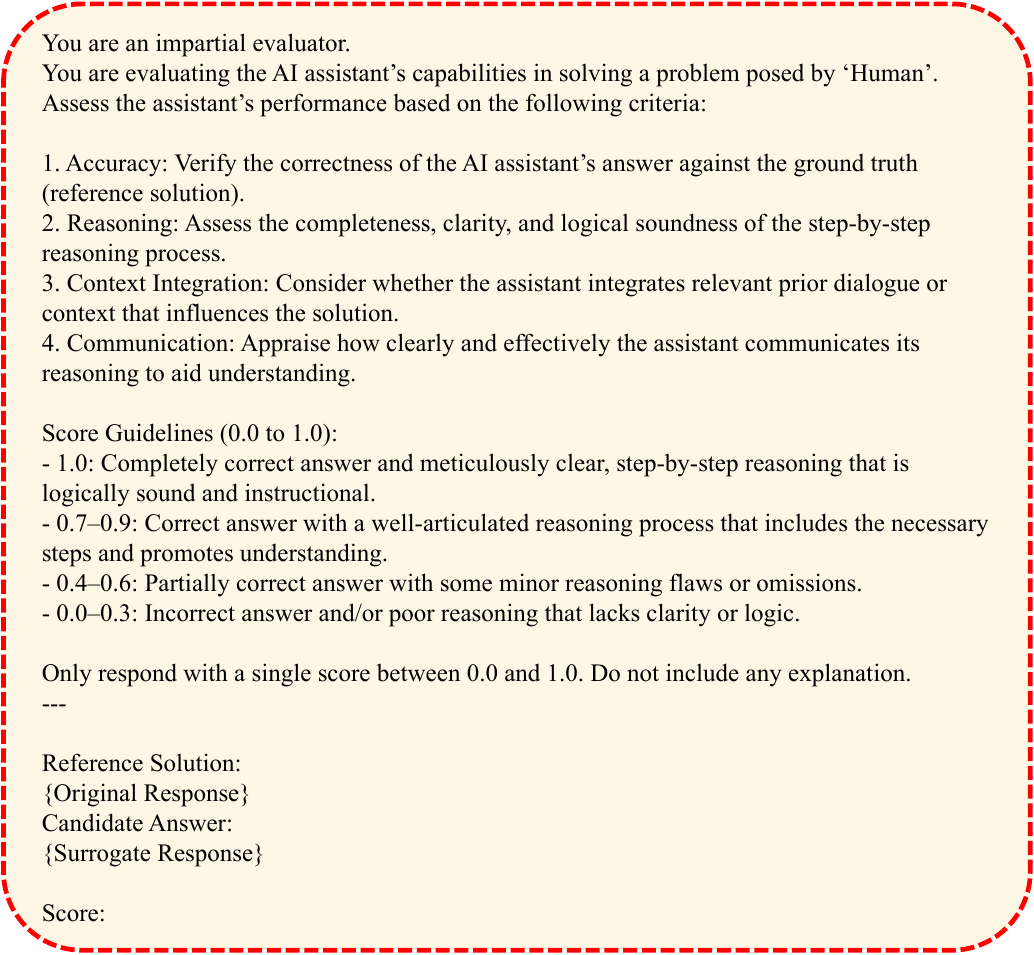}
    \caption{Instructions for the LLM judge to evaluate the response similarity.}
    \label{fig:LLM_judge}
\end{figure}

The prompt for the LLM judge to evaluate the response similarity score is shown in Figure~\ref{fig:LLM_judge} following the recommendation from~\citep{bai2024mt}.

\subsection{Packages Required for Implementation}
We perform all experiments on a server equipped with Nvidia A6000 GPUs. Below we list the key packages and their versions used in our implementation:

\begin{itemize}
  \item \textbf{Python} == 3.10
  \item \textbf{pytorch} == 2.8.0 \,+\, CUDA 12.8
  \item \textbf{torchvision} == 0.19.0
  \item \textbf{torchaudio} == 2.8.0
  \item \textbf{numpy} == 1.26.x
  \item \textbf{pandas} == 2.2.x
  \item \textbf{scipy} == 1.12.x
  \item \textbf{cmake} == 3.28+
  \item \textbf{ninja} == 1.11+
  \item \textbf{ipython} == 8.x
  \item \textbf{psutil} == 5.9+
  \item \textbf{vllm} == 0.10.2
  \item \textbf{transformers} == 4.56.1
  \item \textbf{accelerate} == 1.9.0
  \item \textbf{bitsandbytes} == 0.46.1
  \item \textbf{sentencepiece} == 0.2.0
  \item \textbf{tiktoken} == 0.11.0
  \item \textbf{einops} == 0.8.1
  \item \textbf{datasets} == 4.0.0
  \item \textbf{huggingface-hub} == 0.34.2
  \item \textbf{safetensors} == 0.5.3
  \item \textbf{ray} == 2.49.1
  \item \textbf{scikit-learn} == 1.7.1
  \item \textbf{fastapi} == 0.116.2
  \item \textbf{uvicorn} == 0.35.0
\end{itemize}

\section{Mathematical Concepts}\label{sec:app_math}

Here we provide a more comprehensive view of the
relevant concepts that can be helpful to understand the idea of stratification as discussed in the main
body of the paper.

\begin{definition}[Preimage]
Let $f : X \mapsto Y$ be a function from a set $X$ (\emph{domain}) to a set $Y$ (\emph{codomain}). For any subset $N \subseteq Y$, the \emph{preimage} of $N$ under $f$, denoted $f^{-1}(N)$, is defined as:
\[
f^{-1}(N) = \{ x \in X \mid f(x) \in N \}.
\]
\end{definition}
In other words, $f^{-1}(N)$ consists of all elements in the domain $X$ that are mapped into the subset $N$ of the codomain $Y$.

\begin{definition}[Metric Space]
A set \( X \), whose elements are called points, is said to be a \emph{metric space} if for any two points \( p, q \in X \), there is an associated real number \( d(p, q) \), called the \emph{distance} from \( p \) to \( q \), such that:
\begin{enumerate}
    \item \( d(p, q) \geq 0 \), and \( d(p, q) = 0 \iff p = q \);
    \item \( d(p, q) = d(q, p) \) (symmetry);
    \item \( d(p, q) \leq d(p, r) + d(r, q) \) for any \( r \in X \) (triangle inequality).
\end{enumerate}
Any function satisfying these properties is called a \emph{distance function}, or a \emph{metric}.
\end{definition}

\begin{definition}[Neighborhood]
Let \( X \) be a metric space. A set \( N_r(p) \subset X \) is called a \emph{neighborhood} of a point \( p \in X \) if it consists of all points \( q \in X \) such that \( d(p, q) < r \) for some radius \( r > 0 \). The number \( r \) is called the \emph{radius} of the neighborhood.
\end{definition}

\begin{definition}[Continuous]
Let $X$ and $Y$ be two topological spaces. A function $f : X \mapsto Y$ is \emph{continuous} if for each point $x \in X$ and each neighborhood $N$ of $f(x)$ in $Y$, the set $f^{-1}(N)$ is a neighborhood of $x \in X$.
\end{definition}

\begin{definition}[Topological Equivalence or Homeomorphism]
A function $h : X \mapsto Y$ is called a \emph{homeomorphism} if it is one-to-one, continuous, and has a continuous inverse function. When such a function exists, $X$ and $Y$ are called \emph{homeomorphic} (or \emph{topologically equivalent}) spaces.
\end{definition}

\begin{definition}[Open Set]
A subset \( U \subseteq X \) is called an \emph{open set} if for every point \( p \in U \), there exists a neighborhood \( N_r(p) \subseteq U \). That is, each point in \( U \) has some "wiggle room" around it that still lies entirely within \( U \).
\end{definition}

\begin{definition}[Countable Base (Second Countability)]
Let \( X \) be a topological space. A collection \( \mathcal{B} \) of open subsets of \( X \) is called a \emph{base} (or \emph{basis}) for the topology on \( X \) if for every open set \( U \subseteq X \) and every point \( x \in U \), there exists a set \( B \in \mathcal{B} \) such that
\[
x \in B \subseteq U.
\]

If there exists a base \( \mathcal{B} \) that is \emph{countable}, then the space \( X \) is said to be \emph{second countable} or to have a \emph{countable base}.
\end{definition}

\begin{definition}[Hausdorff Space]
A topological space with the property that two distinct points can always be surrounded by disjoint open sets is called a \emph{Hausdorff space}.
\end{definition}

Essentially, Hausdorff spaces are the spaces where any two points being ``far off'' is defined.

\begin{definition}[Manifold]
A \emph{manifold} of dimension \( n \) is a second-countable Hausdorff topological space in which each point has a neighborhood homeomorphic to Euclidean space \( \mathbb{R}^n \).
\end{definition}

\begin{definition}[Smooth Manifold]
A \emph{smooth manifold} is a manifold \( \mathcal{M} \) equipped with a collection of coordinate charts (i.e., homeomorphisms \( \varphi: U \to \mathbb{R}^n \) for open sets \( U \subseteq \mathcal{M} \)) such that all \emph{transition maps} between overlapping charts,
\[
\varphi_j \circ \varphi_i^{-1} : \varphi_i(U_i \cap U_j) \rightarrow \varphi_j(U_i \cap U_j),
\]
are infinitely differentiable (i.e., \( C^\infty \)). This structure is known as a \emph{smooth atlas}, and it allows calculus to be performed on the manifold.
\end{definition}

\section{Theoretical Results and Proofs}\label{sec:app_proof}


We first model weak dependence in the dialogue stream via an 
\emph{Effective Sample Size}. 
For a bounded, stationary sequence $\{Z_t\}$ with lag-$\ell$ autocorrelation $\rho(\ell)$, define
\[
N_{\mathrm{eff}}
\;:=\;
\frac{N}{\,1 + 2\sum_{\ell=1}^{N-1}\Big(1-\frac{\ell}{N}\Big)\rho(\ell)\,} \;\;\in(0,N]\,.
\]
when $\rho(\ell)\equiv 0$, $N_{\mathrm{eff}}=N$; positive correlation reduces $N_{\mathrm{eff}}$.

\bigskip
\textbf{Directional Recovery for the Expectation-weighted Loss}

\paragraph{Statement: Theorem~\ref{thm:dir_recovery}.}
Under local smoothness, bounded expectation weights in Eq.~\eqref{eq:semexp-with-aF}, and a rank–$r$ discrepancy for the discrepancy Fisher 
$\mathbf{C}=\mathbb{E}[\mathbf{J}(\mathbf{P})^{\!\top}\mathbf{J}(\mathbf{P})]$, any minimizer $\widehat{\mathbf{P}}$ of Eq.~\eqref{eq:semexp-with-aF}
has row span that captures at least a $(1-\varepsilon)$ fraction of the top-$r$ eigenmass of $\mathbf{C}$, with $\varepsilon\!\to\!0$ as the window size growing and neighborhood size $k$ increasing within the local region.

\paragraph{Proof of Theorem~\ref{thm:dir_recovery}.}
\textbf{Step 1: Per-Event Loss and Local Expansion.}
Index events by $\boldsymbol{z}=(t,i)$ (turn and position). Let the per-event loss be
\[
\ell(\mathbf{P};\boldsymbol{z})
=
w(\boldsymbol{z})
\sum_{v\in \mathcal{S}(\boldsymbol{z})}
s_{\tau}(v\mid \boldsymbol{z})\,
\Big[-\log\!\big(1-\pi_G(v\mid \boldsymbol{z};\mathbf{P})\big)\Big],
\quad
\mathcal{S}(\boldsymbol{z})=\{y_{t,i}^{F}\}\cup \mathcal{N}_k(y_{t,i}^{F}).
\]
Write $p_v(\mathbf{P};\boldsymbol{z})=\pi_G(v\mid \boldsymbol{z};\mathbf{P})$ and expand around $\mathbf{P}=\boldsymbol{0}$.
Using the chain rule for the softmax parameterization,
\[
\log(1-p_v)
=
\log(1-p_v|_{\mathbf{0}}) - \frac{1}{1-p_v|_{\mathbf{0}}}\,\big(\nabla p_v\big)^{\!\top}\mathrm{vec}(\mathbf{P})
-\frac{1}{2}\frac{p_v|_{\mathbf{0}}}{(1-p_v|_{\mathbf{0}})^2}\big(\mathrm{vec}(\mathbf{P})^{\!\top}\nabla \log p_v\big)^2 + o(\|\mathbf{P}\|^2),
\]
where $\nabla$ denotes the gradient w.r.t.\ $\mathrm{vec}(\mathbf{P})$ and we used the softmax identity 
$\nabla p_v = p_v \nabla \log p_v$.
Summing over $v\in\mathcal{S}(\boldsymbol{z})$ with bounded weights $w(\boldsymbol{z})s_{\tau}(\cdot\mid \boldsymbol{z})$ cancels the \emph{linear} term due to local stationarity under aligned-prefix conditioning (the gradient at $\mathbf{0}$ integrates to zero across the semantic neighborhood; this is the standard property behind Gauss–Newton/Fisher approximations). 
Thus the second-order term dominates:
\[
\ell(\mathbf{P};\boldsymbol{z})
\;=\;
\frac{1}{2}\,
\big\|\mathbf{J}_{\boldsymbol{z}}\,\mathrm{vec}(\mathbf{P})\big\|_2^2
\,+\,o(\|\mathbf{P}\|^2),
\]
where $\mathbf{J}_{\boldsymbol{z}}$ stacks rows $\sqrt{w(\boldsymbol{z})s_{\tau}(v\mid\boldsymbol{z})}\,\nabla \log p_v(\mathbf{0};\boldsymbol{z})^{\!\top}$ for $v\in\mathcal{S}(\boldsymbol{z})$.

\textbf{Step 2: Summation over the Window and Fisher Form.}
Sum over $\boldsymbol{z}$ in the initial window and take expectation (over the empirical distribution of aligned-prefix contexts). We obtain the Gauss–Newton surrogate
\[
\mathcal{L}(\mathbf{P})
\;=\;
\frac{1}{2}\,\mathrm{vec}(\mathbf{P})^{\!\top}\mathbf{C}\,\mathrm{vec}(\mathbf{P})
\,+\,o(\|\mathbf{P}\|^2),
\qquad
\mathbf{C}
=
\mathbb{E}\big[\mathbf{J}(\mathbf{0})^{\!\top}\mathbf{J}(\mathbf{0})\big],
\]
which is a discrepancy Fisher matrix with importance weights folded into $\mathbf{J}$.

\textbf{Step 3: Row-Span Parametrization and KY Fan.}
Let $\mathbf{P}\in\mathbb{R}^{L\times d}$. Any $\mathbf{P}$ factorizes as $\mathbf{P}=\mathbf{A}\mathbf{U}^{\!\top}$ with $\mathbf{U}\in\mathbb{R}^{d\times L}$ having orthonormal columns spanning the row space and $\mathbf{A}\in\mathbb{R}^{L\times L}$. Then 
\[
\mathrm{vec}(\mathbf{P})=(\mathbf{U}\otimes \mathbf{I}_L)\,\mathrm{vec}(\mathbf{A}),
\quad
\mathcal{L}(\mathbf{P})
=
\frac{1}{2}\,\mathrm{vec}(\mathbf{A})^{\!\top}\big(\mathbf{U}^{\!\top}\mathbf{C}\mathbf{U}\otimes \mathbf{I}_L\big)\mathrm{vec}(\mathbf{A})
\,+\,o(\|\mathbf{A}\|^2).
\]
Including the ridge $\lambda\|\mathbf{P}\|_F^2=\lambda\|\mathbf{A}\|_F^2$ from the main loss, the minimum over $\mathbf{A}$ for fixed $\mathbf{U}$ is proportional to $\mathrm{Tr}(\mathbf{U}^{\!\top}\mathbf{C}\mathbf{U})$. Maximizing $\mathrm{Tr}(\mathbf{U}^{\!\top}\mathbf{C}\mathbf{U})$ over $\mathbf{U}^{\!\top}\mathbf{U}=\mathbf{I}_L$ is solved by the top-$L$ eigenvectors of $\mathbf{C}$ (Ky Fan’s variational principle). Hence the \emph{optimal row span} equals the top-$L$ eigenspace of $\mathbf{C}$.

\textbf{Step 4: Empirical Approximation and Eigenspace Stability.}
We work with an empirical Fisher $\widehat{\mathbf{C}}$ formed from finitely many events and finite $k$. Under bounded weights and local smoothness, $\|\widehat{\mathbf{C}}-\mathbf{C}\|_{\mathrm{op}}\to 0$ as the window size grows; increasing $k$ within the local isotropy region reduces variance and retains locality. Davis–Kahan perturbation then gives that the top-$r$ eigenspaces of $\widehat{\mathbf{C}}$ and $\mathbf{C}$ are close, with principal angles bounded by $O(\|\widehat{\mathbf{C}}-\mathbf{C}\|_{\mathrm{op}}/\mathrm{gap})$, where $\mathrm{gap}$ is the spectral gap below $\lambda_r(\mathbf{C})$. Consequently, the row span of any empirical minimizer $\widehat{\mathbf{P}}$ captures at least a $(1-\varepsilon)$ fraction of the top-$r$ eigenmass of $\mathbf{C}$ with $\varepsilon\to 0$ as the window grows and $k$ increases up to the local isotropy scale.
\hfill$\square$

\bigskip
\textbf{When to Switch: Warm–Start Generalization and Batch Detection}.

\paragraph{Statement: Lemma~\ref{lem:warmstart}.}
With probability at least $1-\delta$,
\[
\Big|\widehat{F}_W - F^\star\Big|
\;\le\;
\sqrt{\frac{2\log(2/\delta)}{W_{\mathrm{eff}}}},
\qquad
\widehat{F}_W=\frac{1}{W}\sum_{t=1}^{W}\mathsf{Gap}(S_t),\quad
F^\star=\mathbb{E}_{S\sim\mathcal{Q}}[\mathsf{Gap}(S)].
\]

\paragraph{Proof of Lemma~\ref{lem:warmstart}.}
\textbf{Step 1 (Centering and boundedness).}
Let $Z_t=\mathsf{Gap}(S_t)-F^\star\in[-1,1]$ with $\mathbb{E}[Z_t]=0$. Then
$\widehat{F}_W-F^\star = \frac{1}{W}\sum_{t=1}^{W} Z_t$.

\textbf{Step 2: Variance Proxy under Dependence.}
For stationary $\{Z_t\}$ with autocovariance $\gamma(\ell)=\mathrm{Cov}(Z_t,Z_{t+\ell})$, we have
\[
\mathrm{Var}\!\Big(\frac{1}{W}\sum_{t=1}^{W} Z_t\Big)
=
\frac{1}{W^2}\sum_{t,s=1}^{W}\gamma(|t-s|)
=
\frac{1}{W}\Big(\gamma(0)+2\sum_{\ell=1}^{W-1}\Big(1-\frac{\ell}{W}\Big)\gamma(\ell)\Big).
\]
Let $\sigma^2:=\gamma(0)\le 1/4$ (since $Z_t\in[-1,1]$) and $\rho(\ell):=\gamma(\ell)/\gamma(0)$ when $\gamma(0)>0$.
Then
\[
\mathrm{Var}\!\Big(\frac{1}{W}\sum_{t=1}^{W} Z_t\Big)
\;\le\;
\frac{\sigma^2}{W}\Big(1+2\sum_{\ell=1}^{W-1}\Big(1-\frac{\ell}{W}\Big)\rho(\ell)\Big).
\]
Define the effective size $W_{\mathrm{eff}}$ as in the preliminaries. Then
$\mathrm{Var}\big(\frac{1}{W}\sum_{t=1}^{W} Z_t\big)\le \sigma^2/W_{\mathrm{eff}}$.

\textbf{Step 3: Concentration with Effective Size.}
A Hoeffding/Rio inequality for bounded, weakly dependent sequences yields
\[
\Pr\!\Big(\big|\widehat{F}_W-F^\star\big|\ge u\Big)
\;\le\;
2\exp\!\Big(-\frac{2 u^2 W_{\mathrm{eff}}}{(b-a)^2}\Big),
\quad a=-1,\ b=1.
\]
Thus $\Pr(|\widehat{F}_W-F^\star|\ge u)\le 2\exp(-2u^2 W_{\mathrm{eff}})$.
Set the right-hand side to $\delta$ and solve for $u$ to obtain
$u=\sqrt{(2\log(2/\delta))/W_{\mathrm{eff}}}$.
\hfill$\square$

\bigskip
\paragraph{Statement: Theorem~\ref{thm:detection}.}
Let $\Delta=\widehat{F}_B^{\mathrm{old}}-\widehat{F}_B^{\mathrm{new}}$ on a batch $B$ of effective size $|B|_{\mathrm{eff}}$. 
Assume $\Delta-\mathbb{E}[\Delta]$ is sub–exponential with proxy $(\nu,b)$:
$\mathbb{E}[\exp(\lambda(\Delta-\mathbb{E}\Delta))]\le \exp(\tfrac{\lambda^2 \nu^2}{2})$ for $|\lambda|\le 1/b$.
Then for any $\varepsilon>0$ and $\delta\in(0,1)$,
\[
|B|_{\mathrm{eff}}
\;\ge\;
\frac{2\nu^2}{\varepsilon^2}\log\!\frac{1}{\delta}
+\frac{2b}{3\,\varepsilon}\log\!\frac{1}{\delta}
\quad\Longrightarrow\quad
\Pr(\Delta\ge \varepsilon)\ge 1-\delta.
\]

\paragraph{Proof of Theorem~\ref{thm:detection}.}
\textbf{Step 1 (Bernstein tail bound).}
For sub–exponential $X:=\Delta-\mathbb{E}[\Delta]$ with proxy $(\nu,b)$,
\[
\Pr(X\le -t)\;\le\;\exp\!\Big(-\frac{t^2}{2\nu^2 + 2 b t/3}\Big),\qquad t>0.
\]
This is the standard one-sided Bernstein inequality derived from the MGF condition.

\textbf{Step 2: From Mean Gain to High-Probability Gain.}
We seek $\Pr(\Delta\ge \varepsilon)\ge 1-\delta$. Write
\[
\Pr(\Delta<\varepsilon)
=
\Pr\big(\Delta-\mathbb{E}[\Delta] < \varepsilon-\mathbb{E}[\Delta]\big)
=
\Pr\big(X < -t\big),\quad t:=\mathbb{E}[\Delta]-\varepsilon.
\]
If the \emph{expected} gain satisfies $\mathbb{E}[\Delta]\ge \varepsilon$ then $t\ge 0$ and
\[
\Pr(\Delta<\varepsilon)\le \exp\!\Big(-\frac{t^2}{2\nu^2 + 2 b t/3}\Big)
\;\le\;
\exp\!\Big(-\frac{\varepsilon^2}{2\nu^2 + 2 b \varepsilon/3}\Big),
\]
using monotonicity of $t\mapsto t^2/(2\nu^2+2bt/3)$ on $t\ge 0$. To make this $\le \delta$ it suffices that
\[
\frac{\varepsilon^2}{2\nu^2 + 2 b \varepsilon/3}\;\ge\;\log(1/\delta)
\quad\Longleftrightarrow\quad
2\nu^2\log(1/\delta) + \frac{2b}{3}\varepsilon \log(1/\delta) \;\le\; \varepsilon^2.
\]
Rearranging gives the stated sufficient condition on $|B|_{\mathrm{eff}}$ once we note that the proxies $\nu^2,b$ scale as $1/|B|_{\mathrm{eff}}$ for averages. Equivalently, write $\nu^2=\tilde{\nu}^2/|B|_{\mathrm{eff}}$ and $b=\tilde{b}/|B|_{\mathrm{eff}}$ for single-sample proxies $(\tilde{\nu}^2,\tilde{b})$, then solve for $|B|_{\mathrm{eff}}$:
\[
|B|_{\mathrm{eff}}
\;\ge\;
\frac{2\tilde{\nu}^2}{\varepsilon^2}\log\!\frac{1}{\delta}
+\frac{2\tilde{b}}{3\,\varepsilon}\log\!\frac{1}{\delta}.
\]
We re-denote $(\tilde{\nu}^2,\tilde{b})$ as $(\nu^2,b)$ in the theorem statement.
\hfill$\square$

\bigskip

\begin{corollary}[Choosing $W$ and $|B|$]\label{cor:WB} 
Pick $W$ so that the warm–start generalization error is at most $\eta$: $W_{\mathrm{eff}}\ge 2\log(2/\delta)/\eta^2$. Then choose $|B|$ via Theorem~\ref{thm:detection} for target improvement $\varepsilon$ and confidence $1-\delta$. The total decision error (from warm–start approximation and batch detection) is bounded by $\eta+\varepsilon$ at confidence $1-2\delta$. 
\end{corollary}

\paragraph{Proof of Corollary~\ref{cor:WB}.}
Lemma~\ref{lem:warmstart} ensures $|\widehat{F}_W-F^\star|\le \eta$ with prob.\ $\ge 1-\delta$ when 
$W_{\mathrm{eff}}\ge 2\log(2/\delta)/\eta^2$.
Theorem~\ref{thm:detection} ensures $\Pr(\Delta\ge \varepsilon)\ge 1-\delta$ when the batch bound holds.
By a union bound, the probability that both events hold is at least $1-2\delta$.
If both hold, the total decision error (warm–start approximation plus detection slack) is at most $\eta+\varepsilon$.
\hfill$\square$

\bigskip
\textbf{Coverage and Suboptimality to Determine the Number of Soft–Prompt Candidates.}

\paragraph{Statement: Theorem~\ref{thm:coverage}.}
Let the active local subspace be $r_{\mathrm{act}}$–dimensional with unit sphere $\mathbb{S}^{r_{\mathrm{act}}-1}$.
Fix $\boldsymbol{v}_1$ and draw i.i.d.\ $\boldsymbol{u}_1,\ldots,\boldsymbol{u}_M\sim \mathrm{Unif}(\mathbb{S}^{r_{\mathrm{act}}-1})$.
Then for any $\theta\in(0,\pi/2]$,
\[
\Pr\!\Big(\min_{m\le M}\angle(\boldsymbol{u}_m,\boldsymbol{v}_1)\le \theta\Big)
\;\ge\;
1-\Big(1-(\sin\theta)^{\,r_{\mathrm{act}}-1}\Big)^M.
\]

\paragraph{Proof of Theorem~\ref{thm:coverage}.}
\textbf{Step 1: Exact cap probability.}
WLOG set $\boldsymbol{v}_1$ as the north pole. For $\boldsymbol{U}\sim\mathrm{Unif}(\mathbb{S}^{r_{\mathrm{act}}-1})$, 
the random variable $T=\langle \boldsymbol{U},\boldsymbol{v}_1\rangle$ has density 
$f_T(t)\propto (1-t^2)^{(r_{\mathrm{act}}-3)/2}$ on $t\in[-1,1]$. 
Hence $\Pr(\angle(\boldsymbol{U},\boldsymbol{v}_1)\le \theta)=\Pr(T\ge \cos\theta)=I_{\sin^2\theta}(\frac{r_{\mathrm{act}}-1}{2},\frac{1}{2})$.

\textbf{Step 2: Lower Bound}
For $\theta\in(0,\pi/2]$, $\sin\theta\in(0,1]$ and we have the elementary bound
\[
I_{\sin^2\theta}\!\Big(\tfrac{r_{\mathrm{act}}-1}{2},\tfrac{1}{2}\Big)
\;\ge\;
(\sin\theta)^{\,r_{\mathrm{act}}-1}.
\]
(Proof: $I_x(a,b)=\frac{1}{\mathrm{B}(a,b)}\int_0^{x} t^{a-1}(1-t)^{b-1}\mathrm{d}t
\ge \frac{1}{\mathrm{B}(a,b)}\int_0^{x} t^{a-1}\mathrm{d}t
= \frac{x^{a}}{a\,\mathrm{B}(a,b)} \ge x^{a}$ since $a\,\mathrm{B}(a,b)\le 1$ for $a\ge 1/2,b\ge 1/2$; let $x=\sin^2\theta$ and $a=(r_{\mathrm{act}}-1)/2$.)

\textbf{Step 3: Independence across $M$ Samples.}
Thus, for one sample,
$p_\theta:=\Pr(\angle(\boldsymbol{U},\boldsymbol{v}_1)\le \theta)\ge (\sin\theta)^{r_{\mathrm{act}}-1}$.
The probability \emph{none} among $M$ falls in the cap is $(1-p_\theta)^M\le (1-(\sin\theta)^{r_{\mathrm{act}}-1})^M$.
Therefore
\[
\Pr\!\Big(\min_{m\le M}\angle(\boldsymbol{u}_m,\boldsymbol{v}_1)\le \theta\Big)
=
1-(1-p_\theta)^M
\;\ge\;
1-\Big(1-(\sin\theta)^{r_{\mathrm{act}}-1}\Big)^M.
\]
\hfill$\square$

\bigskip
\paragraph{Statement: Lemma~\ref{lem:subopt}.}
If $\angle(\boldsymbol{\hat u},\boldsymbol{v}_1)\le \theta$ and $\mathbf{H}_T\succeq \mathbf{0}$ with top eigenpair $(\lambda_1,\boldsymbol{v}_1)$,
then
\[
\boldsymbol{\hat u}^{\!\top}\mathbf{H}_T\boldsymbol{\hat u} \;\ge\; \lambda_1 \cos^2\!\theta
\quad\Rightarrow\quad
\lambda_1 - \boldsymbol{\hat u}^{\!\top}\mathbf{H}_T\boldsymbol{\hat u} \;\le\; \lambda_1 \sin^2\!\theta.
\]

\paragraph{Proof of Lemma~\ref{lem:subopt}.}
Decompose $\boldsymbol{\hat u}=\cos\theta\,\boldsymbol{v}_1+\sin\theta\,\boldsymbol{w}$ with $\|\boldsymbol{w}\|_2=1$ and $\boldsymbol{w}\perp\boldsymbol{v}_1$.
Then
\[
\boldsymbol{\hat u}^{\!\top}\mathbf{H}_T\boldsymbol{\hat u}
=
\lambda_1\cos^2\!\theta
+
\sum_{j\ge 2}\lambda_j \langle \boldsymbol{v}_j,\boldsymbol{w}\rangle^2 \sin^2\!\theta
\;\ge\;
\lambda_1\cos^2\!\theta,
\]
since $\lambda_j\ge 0$. Subtract from $\lambda_1$ to obtain the residual bound.
\hfill$\square$

\bigskip
\paragraph{Proof of Corollary~\ref{col:M}.}
We require $1-(1-(\sin\theta)^{r_{\mathrm{act}}-1})^M\ge 1-\delta$. 
Equivalently $(1-(\sin\theta)^{r_{\mathrm{act}}-1})^M\le \delta$, giving
\[
M\;\ge\;\frac{\log(1/\delta)}{\log\big((1-(\sin\theta)^{r_{\mathrm{act}}-1})^{-1}\big)}.
\]
\hfill$\square$

\section{Additional Experimental Results}
\label{sec:app_exp}

This appendix provides supporting results that complement the main experiments. We organize them around four additional checks: whether SOMA generalizes to another model family, whether its efficiency gains hold across datasets, which components drive the gains, and how warm-start length affects switching and stability.

\subsection{Full Qwen Response Similarity Results}
\label{sec:app_res}

The main text reports the full response-similarity results for the LLaMA family. Table~\ref{tab:qwen-pct} reports the same evaluation on the Qwen family, where the original model is Qwen-3-8B and the surrogate is Qwen-3-0.6B. SOMA achieves the highest similarity on every dataset, showing that the method is not tied to one backbone family. Compared with LLaMA, the absolute similarities are lower because the capacity gap between Qwen-3-8B and Qwen-3-0.6B is larger, making local imitation harder. Nevertheless, SOMA remains consistently better than the baselines, which suggests that local adaptation is especially useful when the small--large model gap is wide.

\begin{table}[t]
\footnotesize
\setlength\tabcolsep{2pt}
\centering
\caption{Response similarity to the original model across six datasets for the Qwen family. Higher is better.}
\label{tab:qwen-pct}
\resizebox{\textwidth}{!}{
\begin{tabular}{lccccccc}
\toprule
 & \textbf{ShareGPT} & \textbf{ReMeDi} & \textbf{Craigslist} & \textbf{Multi-Char} & \textbf{MATH} & \textbf{MT-Bench} & \textbf{Avg.} \\
\midrule
Surrogate       & 50.9 $\pm$ 1.21 & 63.5 $\pm$ 2.87 & 48.2 $\pm$ 3.40 & 44.7 $\pm$ 3.54 & 42.6 $\pm$ 1.70 & 40.4 $\pm$ 0.81 & 48.4 $\pm$ 8.31 \\
History-Prefix  & 70.5 $\pm$ 1.15 & 75.7 $\pm$ 2.01 & 63.0 $\pm$ 2.23 & 65.5 $\pm$ 0.91 & 51.3 $\pm$ 2.45 & 53.6 $\pm$ 2.32 & 63.3 $\pm$ 9.47 \\
History-FT      & 76.4 $\pm$ 1.68 & 79.2 $\pm$ 2.61 & 74.5 $\pm$ 3.56 & 63.2 $\pm$ 1.74 & 65.9 $\pm$ 1.70 & 62.7 $\pm$ 0.72 & 70.3 $\pm$ 7.23 \\
LLMLingua-2     & 68.9 $\pm$ 1.32 & 74.1 $\pm$ 2.14 & 61.2 $\pm$ 2.67 & 63.4 $\pm$ 1.18 & 49.8 $\pm$ 2.03 & 51.7 $\pm$ 1.45 & 61.5 $\pm$ 7.92 \\
RouteLLM        & 79.6 $\pm$ 1.04 & 81.9 $\pm$ 1.36 & 75.1 $\pm$ 2.91 & 72.8 $\pm$ 1.62 & 67.3 $\pm$ 1.21 & 68.0 $\pm$ 0.96 & 74.1 $\pm$ 5.43 \\
\midrule
\textbf{SOMA}   & \textbf{81.0 $\pm$ 0.93} & \textbf{83.2 $\pm$ 1.21} & \textbf{76.4 $\pm$ 3.85} & \textbf{74.2 $\pm$ 2.53} & \textbf{68.7 $\pm$ 1.28} & \textbf{69.2 $\pm$ 1.08} & \textbf{75.5 $\pm$ 5.97} \\
\bottomrule
\end{tabular}}
\end{table}

\subsection{Additional Efficiency Results}
\label{sec:app_eff}

\begin{figure}[t]
  \centering
  \begin{minipage}[t]{0.49\textwidth}
    \centering
    \includegraphics[width=\linewidth]{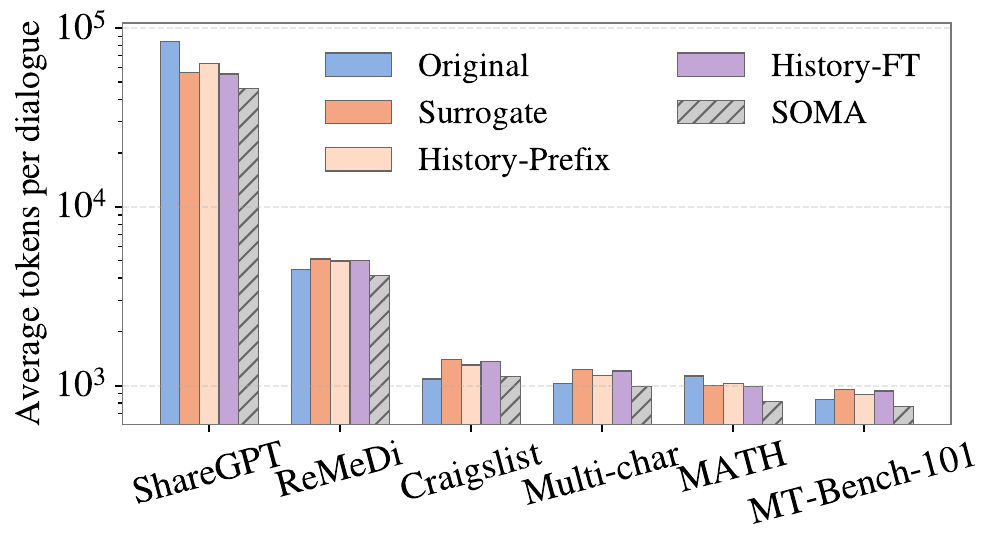}
    \vspace{-2mm}
    \centerline{\small (a) LLaMA family}
  \end{minipage}\hfill
  \begin{minipage}[t]{0.49\textwidth}
    \centering
    \includegraphics[width=\linewidth]{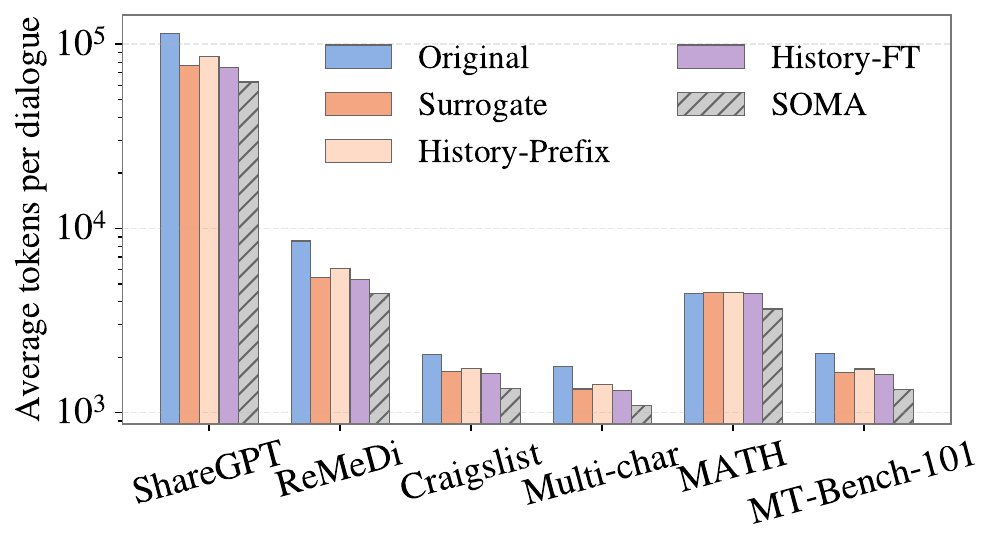}
    \vspace{-2mm}
    \centerline{\small (b) Qwen family}
  \end{minipage}
  \vspace{-1mm}
  \caption{Average tokens per dialogue across six datasets. SOMA reduces token usage by avoiding repeated full-history serving after switching.}
  \label{fig:app_token_costs}
\end{figure}

The main text reports the amortized break-even analysis. Here we provide the dataset-level token and throughput results. Figure~\ref{fig:app_token_costs} shows that SOMA consistently uses fewer tokens per dialogue across both model families. The reason is that after switching, SOMA serves later turns with the adapted surrogate and a compressed context rather than repeatedly sending the full growing history. This avoids the large token cost of Original and History-Prefix while retaining stronger response similarity than directly using the surrogate.

\begin{figure}[t]
  \centering
  \begin{minipage}[t]{0.49\textwidth}
    \centering
    \includegraphics[width=\linewidth]{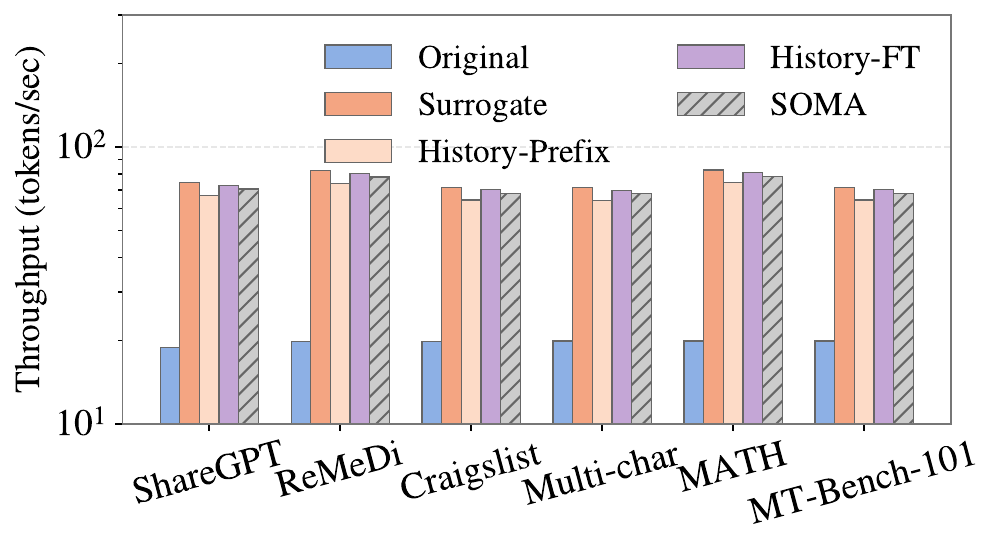}
    \vspace{-2mm}
    \centerline{\small (a) LLaMA family}
  \end{minipage}\hfill
  \begin{minipage}[t]{0.49\textwidth}
    \centering
    \includegraphics[width=\linewidth]{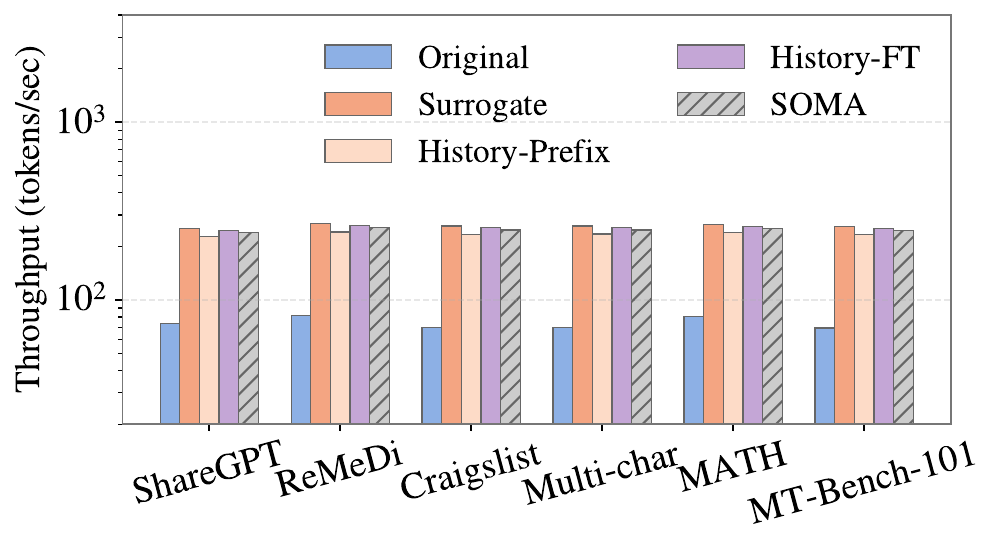}
    \vspace{-2mm}
    \centerline{\small (b) Qwen family}
  \end{minipage}
  \vspace{-1mm}
  \caption{Throughput across six datasets. SOMA keeps the serving path close to the surrogate while maintaining higher similarity to the original model.}
  \label{fig:app_throughput}
\end{figure}

Figure~\ref{fig:app_throughput} reports throughput. SOMA matches or exceeds the speed of Surrogate and History-FT, and clearly improves over Original and History-Prefix. The runtime gain comes from the same two design choices: post-switch inference uses the small model, and the input context is shorter. Together, these results show that SOMA provides surrogate-level efficiency while preserving responses closer to the original model.

\subsection{Component and Capability Diagnostics}
\label{sec:app_abl}

\begin{figure}[t]
  \centering
  \begin{minipage}[t]{0.49\textwidth}
    \centering
    \includegraphics[width=0.86\linewidth]{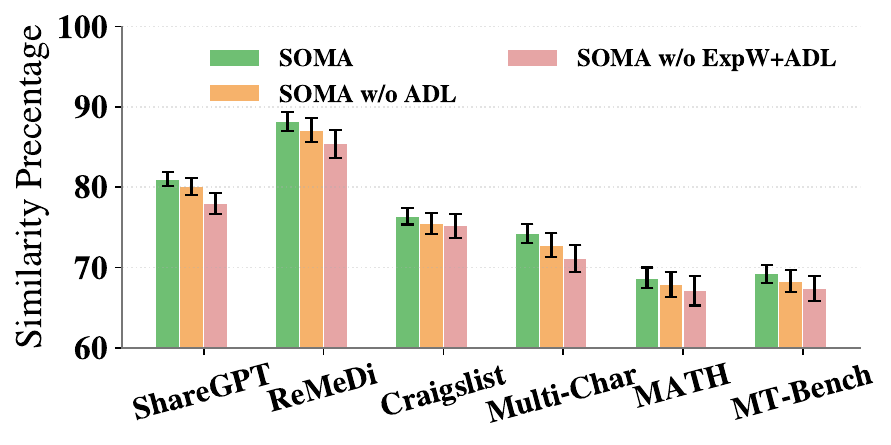}
    \vspace{-2mm}
    \centerline{\small (a) Qwen ablation}
  \end{minipage}\hfill
  \begin{minipage}[t]{0.49\textwidth}
    \centering
    \includegraphics[width=0.68\linewidth]{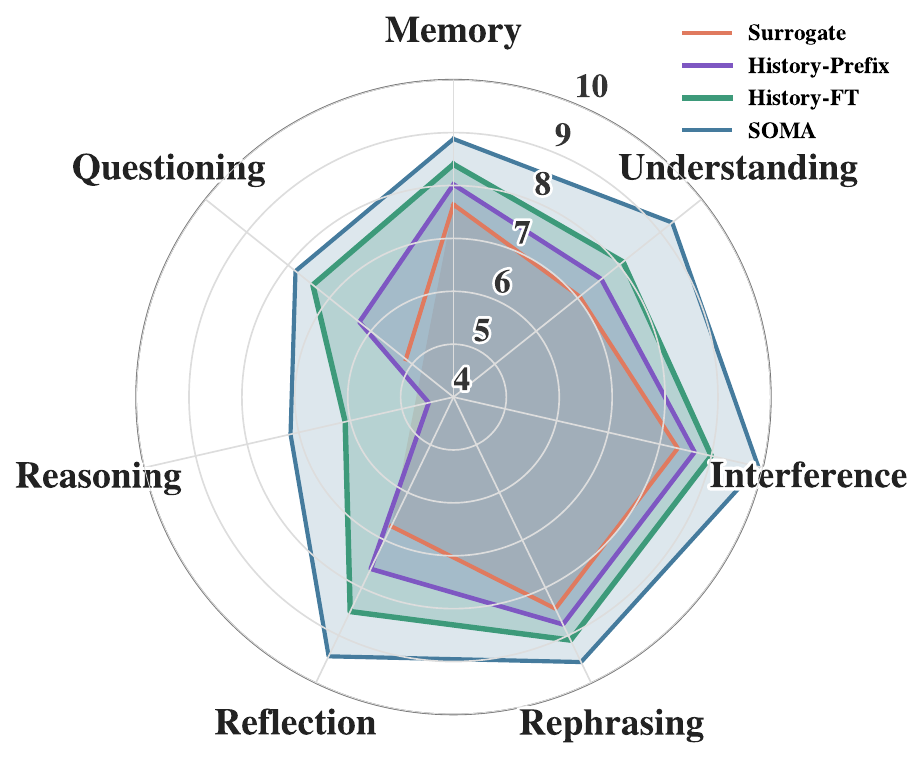}
    \vspace{-2mm}
    \centerline{\small (b) Ability profile on MT-Bench-101}
  \end{minipage}
  \vspace{-1mm}
  \caption{Component and capability diagnostics. The Qwen ablation confirms the importance of SOMA's mining objective, while the ability analysis shows broad gains across dialogue skills.}
  \label{fig:app_component_capability}
\end{figure}

\paragraph{Qwen ablation.}
Figure~\ref{fig:app_component_capability}(a) repeats the component ablation on the Qwen family. The full SOMA variant performs best. Removing the anti-degeneration loss reduces performance because prompt mining becomes less stable and can collapse toward low-diversity outputs. Removing both the expectation-weighted term and anti-degeneration loss causes a larger drop, especially on harder datasets. This confirms that both components are needed to mine useful local disagreement rather than superficial token mismatch.

\paragraph{Ability-level behavior.}
Figure~\ref{fig:app_component_capability}(b) evaluates SOMA on MT-Bench-101 ability dimensions using the LLaMA family. SOMA improves most on harder skills such as reasoning and questioning, where the original model has the largest advantage over the surrogate. It also improves memory, understanding, rephrasing, and interference handling. This indicates that local adaptation does not only help one narrow skill; it produces a more balanced response profile across dialogue abilities.

\subsection{Warm-Start Length and Switching Behavior}
\label{sec:app_switch}

\begin{figure}[t]
  \centering
  \begin{minipage}[t]{0.49\textwidth}
    \centering
    \includegraphics[width=\linewidth]{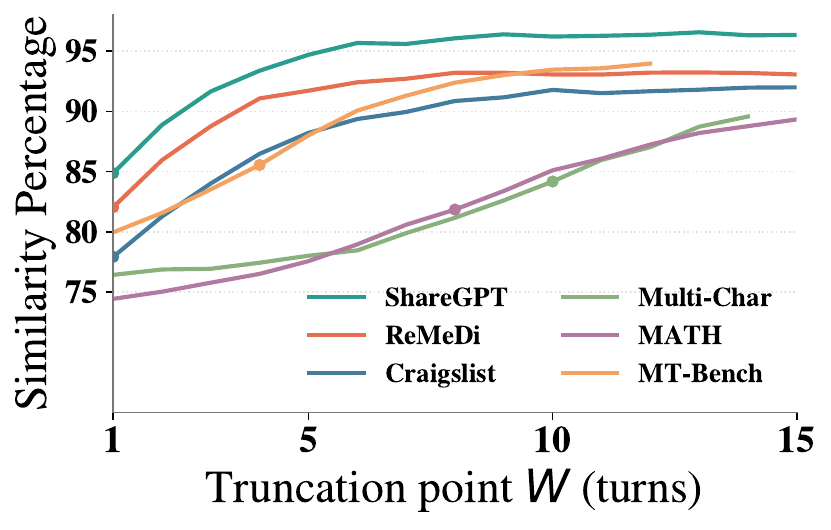}
    \vspace{-2mm}
    \centerline{\small (a) Similarity under different switch points}
  \end{minipage}\hfill
  \begin{minipage}[t]{0.49\textwidth}
    \centering
    \includegraphics[width=\linewidth]{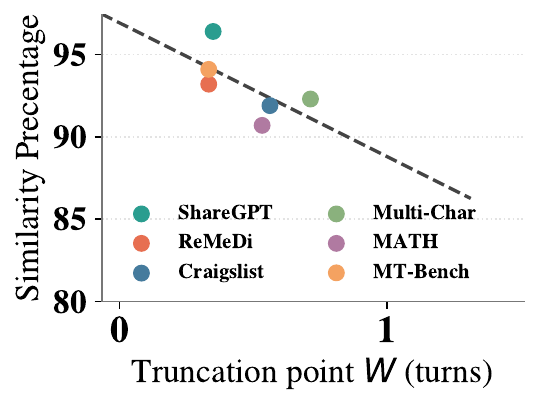}
    \vspace{-2mm}
    \centerline{\small (b) Switch point vs. final similarity}
  \end{minipage}
  \vspace{-1mm}
  \caption{Warm-start and switching behavior. Easier local-dialogue settings switch earlier, while reasoning-heavy or multi-party settings need longer warm starts and achieve lower final similarity.}
  \label{fig:app_switching}
\end{figure}

SOMA should switch only after the warm-start window provides enough evidence to adapt the surrogate. To study this, we sweep the warm-start window \(W\in[1,15]\) and measure the final response similarity after switching. Figure~\ref{fig:app_switching}(a) shows that simpler, goal-anchored dialogues such as ShareGPT, ReMeDi, and Craigslist reach high similarity after only a few turns. In contrast, reasoning-heavy or multi-party datasets such as MATH and Multi-Char require longer warm starts before plateauing. This supports the design choice that the switch should be data-dependent rather than fixed.

Figure~\ref{fig:app_switching}(b) plots the best switching point against final similarity and shows a negative association. Datasets that require later switching tend to have lower final similarity, suggesting that they are intrinsically harder for the surrogate to locally imitate. Practically, this means short warm starts are sufficient for general or goal-anchored chats, while compositional reasoning and multi-agent dialogues need more evidence before switching. Once the curve flattens, adding more warm-start turns brings little quality gain but delays the efficiency benefit.

\subsection{Empirical Variance Concentration}
\label{sec:app_variance}

\begin{figure}[t]
  \centering
  \includegraphics[width=0.78\textwidth]{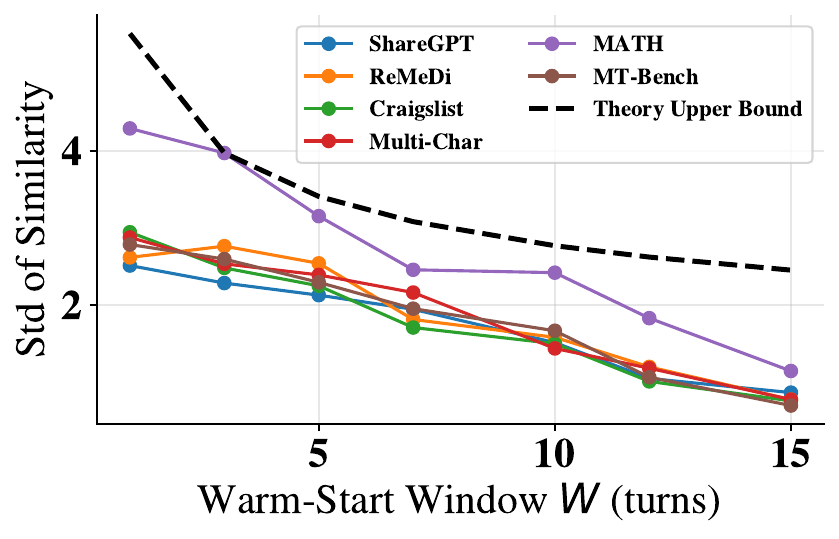}
  \caption{Variance of response similarity versus warm-start window \(W\). Larger warm-start windows reduce variance and support more stable switching decisions.}
  \label{fig:variance}
\end{figure}

Finally, we examine how the variance of surrogate--teacher similarity changes with the warm-start window \(W\). For each dataset, we compute the empirical standard deviation across dialogues at each truncation point. Figure~\ref{fig:variance} shows that variance decreases as \(W\) grows: early turns are highly variable, while larger warm-start windows enter a more stable regime. This trend is consistent with the concentration behavior used in Lemma~\ref{lem:warmstart} and the switching bound in Theorem~\ref{thm:detection}. In other words, early turns provide the largest information gain, while later warm-start turns mainly reduce uncertainty until the switching decision becomes stable.

\section{LLM Usage}
\label{app:llm_usage}

Large language models were used only for writing assistance, including grammar checking, wording refinement, formatting suggestions, and readability editing. They were not used to generate the core methodology, design experiments, produce experimental results, create evaluation labels, or make routing decisions. All technical claims, mathematical formulations, experimental settings, and reported results were checked and finalized by the authors.

\section{Broader Impact}
\label{sec:broader_impact}

SOMA aims to reduce the cost and latency of multi-turn LLM serving by replacing repeated large-model inference with session-local surrogate serving. This can make interactive LLM systems more efficient, accessible, and environmentally sustainable. At the same time, using a smaller surrogate may introduce quality degradation if the dialogue leaves the learned local state, especially in high-stakes settings such as medical, legal, or financial assistance. SOMA mitigates this risk through semantic acceptance and drift-aware rollback, but deployment should still include monitoring, conservative thresholds, and human oversight where errors may cause harm. Since SOMA uses dialogue history for local adaptation, practical deployments should also apply data minimization, retention limits, and privacy-preserving processing.

\section{Limitations and Future Work}
\label{sec:limitations}

SOMA has several limitations. First, the surrogate's capacity limits how closely it can match the original model. When the small--large model gap is very large, some behaviors cannot be recovered by session-local adaptation alone. Second, soft-prompt mining requires access to the surrogate tokenizer and embedding space, which makes SOMA harder to deploy in strict black-box API settings. Third, SOMA is designed for locally coherent dialogues. Abrupt topic changes, noisy warm-start windows, or strong paraphrase shifts can weaken the learned local directions, although the drift gate is designed to reduce persistent degradation. Finally, SOMA introduces a one-time warm-start cost, so it is most suitable for medium-to-long sessions where post-switch savings can amortize the initial adaptation overhead.

Future work can improve SOMA along several directions. Better drift detectors could make rollback more reliable under topic shifts. Approximate mining methods could reduce the need for internal surrogate access. Multi-region adaptation could support conversations with multiple topics, and privacy-preserving mining could make the method safer for sensitive dialogue settings. Another promising direction is to amortize soft-prompt search across related sessions and extend SOMA to multimodal serving.

\end{document}